\documentclass[10pt,journal,compsoc]{IEEEtran}
\usepackage{amsmath,amsfonts,amssymb}
\usepackage{algorithmic}
\usepackage{algorithm}
\usepackage{array}
\usepackage{textcomp}
\usepackage{stfloats}
\usepackage{url}
\usepackage{verbatim}
\usepackage{graphicx}
\usepackage{cite}
\usepackage{booktabs}
\usepackage{multirow}
\usepackage{float}
\usepackage{caption}
\usepackage[dvipsnames,table]{xcolor}
\usepackage{colortbl}
\usepackage[pagebackref=false,breaklinks=true,colorlinks,bookmarks=false,
            citecolor=blue,linkcolor=blue,urlcolor=blue]{hyperref}

\begin{document}

\newcommand{\question}{%
    \stepcounter{question}%
    \textbf{Q\thequestion:~\ignorespaces}%
}
\newcounter{question}
\setcounter{question}{0}

\definecolor{lightblue}{HTML}{f3f7fc}

\title{GIFT: Guided Intermediate Feature Training via Action-Oriented Structural Supervision for Robotic Manipulation}

\author{Yupeng Zheng$^{1,2*}$, Xiang Li$^{3*}$, Songen Gu$^{4*}$, Yuhang Zheng$^{5*}$, Shuai Tian$^{1,2}$, Weize Li$^{5}$, \\
Linbo Wang$^{1,2}$, Chaoyue Li$^{1,2}$, Qichao Zhang$^{1,2\dagger}$, Haoran Li$^{1,2\dagger}$, Zhongpu Xia$^{1}$, \\
Ya-Qin Zhang$^{3}$, Shuicheng Yan$^{5}$, Dongbin Zhao$^{1,2\dagger}$ \\
\textsuperscript{1}Institute of Automation, Chinese Academy of Sciences,
\textsuperscript{2}University of Chinese Academy of Sciences,\\
\textsuperscript{3}Tsinghua University,
\textsuperscript{4}Fudan University,
\textsuperscript{5}National University of Singapore\\
\textsuperscript{*}Equal contribution. \textsuperscript{\textdagger}Corresponding authors.\\
}



\IEEEtitleabstractindextext{%
\begin{abstract}
Vision-language pre-training and predictive world modeling provide robot policies with rich semantic and dynamic visual features, but their native action and visual-prediction objectives may omit critical physical and task structure while retaining control-irrelevant visual redundancy, leading to suboptimal performance.
We call this mismatch between visual richness and control utility the action-sufficiency gap.
We investigate whether this gap can be bridged by guiding intermediate features to preserve three recurring types of control-relevant structure in robotic manipulation: geometry governing motion feasibility, affordance encoding instruction-relevant entities and object-centric end-effector configurations, and goals grounding instructions in task-relevant action regions.
To this end, we present GIFT (Guided Intermediate Feature Training), an architecture-flexible framework for learning intermediate features that translates these structures into training-time constraints through geometry alignment, affordance prediction, and goal-region reconstruction.
We instantiate GIFT in a semantics-centered Vision-Language-Action (VLA) policy, a direct-action World-Action Model (WAM), and an inverse-dynamics WAM, and compare injection and no-injection designs while retaining each model's action formulation.
Under zero-shot transfer to LIBERO-Plus, GIFT-VLA, GIFT-WAM-Fast, and GIFT-WAM-IDM outperform StarVLA-OFT, Fast-WAM, and Fast-WAM-IDM by 4.6, 12.6, and 5.2 percentage points, reaching 79.6\%, 72.6\%, and 87.8\%, respectively, across seven distribution shifts.
On RoboCasa, the three GIFT variants reach 61.4\%, 83.6\%, and 82.3\%, outperforming their counterparts by 12.6, 9.0, and 8.4 points, respectively.
Together, these results establish learning functionally structured intermediate features as a reusable principle across model-specific action formulations, with especially large gains on articulated-object tasks and high-precision real-world manipulation under unseen visual and spatial perturbations.
Ablation studies further show that the no-injection variants perform better, indicating that GIFT internalizes task-relevant structure into intermediate features without relying on auxiliary conditioning. Project page: \url{https://openphoenix-team.github.io/GIFT-pages}.
\end{abstract}

}

\maketitle

\IEEEdisplaynontitleabstractindextext
\IEEEpeerreviewmaketitle

\section{Introduction}

\begin{figure}[H]
\centering
\includegraphics[width=0.72\linewidth]{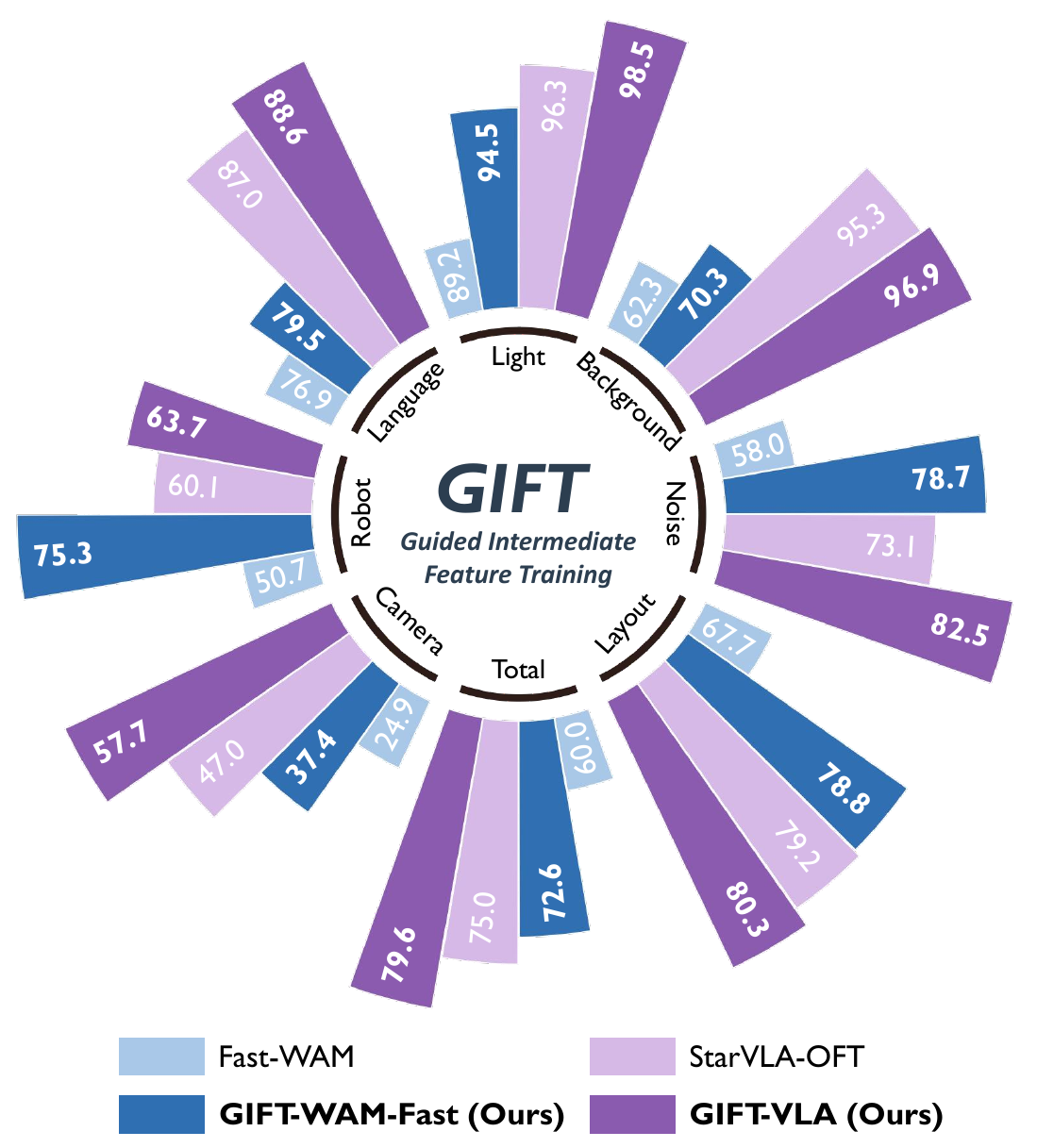}
\caption{\textbf{GIFT improves zero-shot robustness.}
GIFT-VLA and GIFT-WAM-Fast consistently outperform
StarVLA-OFT and Fast-WAM under seven LIBERO-Plus shifts
and on average.}
\label{fig:teaser}
\end{figure}

\IEEEPARstart{G}{eneral-purpose} robot policies must connect semantic task understanding with precise physical control.
Vision-Language-Action (VLA) policies transfer vision-language knowledge to action prediction~\cite{zitkovich2023rt,kim2025openvla,black2024pi0}, while World-Action Models (WAMs) couple control with predictive visual dynamics~\cite{kim2025cosmospolicy,yuan2026fast,ma2026dit4dit}.
Both pursue reusable language-conditioned manipulation, yet matched baselines remain suboptimal under viewpoint, layout, language,
appearance, and robot-initialization shifts (Fig.~\ref{fig:teaser}), revealing a mismatch between visual richness and control utility.

Despite this progress, neither end-to-end action supervision nor visual prediction guarantees that intermediate features retain the information most useful for robot control.
In VLAs, action-only supervision can reward shortcuts based on background textures or camera artifacts and does not explicitly preserve metric relations, object-centric interaction configurations, or instruction-conditioned targets~\cite{jia2026guidedvla}.

In WAMs, RGB latents optimized for pixel reconstruction emphasize appearance; even visually plausible forecasts may retain task-irrelevant detail while omitting the 3D and object-semantic structure needed for interaction~\cite{su2026worldguidance,yan2026flex}.
We call this mismatch between visual richness and control utility the \emph{action-sufficiency gap}.
Recent methods address parts of this gap through auxiliary supervision.
Geometry-oriented approaches inject explicit 3D representations or align intermediate features with geometric teachers~\cite{qu2025spatialvla,sun2025geovla,li2025spatial}, while object- and goal-oriented methods introduce affordance reasoning or reconstruct instruction-relevant regions~\cite{li2025coavlaimprovingvisionlanguageactionmodels,song2026reconvla}.
DreamVLA predicts dynamic, spatial, and semantic knowledge~\cite{zhang2025dreamvla}; Flex-$\pi$ jointly supervises RGB, 3D pointmaps, and object-level semantics~\cite{yan2026flex}; GuidedVLA assigns object, skill, and geometry factors to specialized decoder heads~\cite{jia2026guidedvla}; and World Guidance compresses future observations into an action-conditioning space~\cite{su2026worldguidance}.
Although these methods validate specific guidance signals, their studies are generally confined to selected factors and a particular policy family, attention layout, or predictive interface.
It therefore remains unclear whether a common principle for guiding intermediate features can preserve complementary physical and task structures across diverse policy architectures, and whether structuring these features alone can improve control without injecting auxiliary predictions into the action head.

To investigate this question, our core insight is to develop an architecture-flexible, unified principle for guiding intermediate features to preserve control-relevant structures that recur in robotic manipulation.
Specifically, we organize the guidance around three complementary types of information: \emph{geometry} governing motion feasibility, \emph{affordance} encoding instruction-relevant entities and object-centric end-effector configurations, and \emph{goals} grounding instructions in task-relevant action regions.
Building on this insight, we propose \textbf{GIFT} (\textbf{G}uided \textbf{I}ntermediate \textbf{F}eature \textbf{T}raining), an architecture-flexible framework that translates these structures into training constraints on intermediate visual features through geometry alignment, affordance prediction, and goal-region reconstruction.
The key is not to prescribe a shared action-reasoning interface, but to specify what control-relevant information intermediate features should retain across policies.

To test whether this principle for guiding intermediate features transfers across action formulations, we instantiate GIFT in three policy configurations: a semantics-centered VLA with direct action prediction, a fast-mode WAM with direct action diffusion, and an inverse-dynamics WAM that derives actions from predicted future trajectories.
All three configurations share the same geometry--affordance--goal supervision while retaining their model-specific action formulations.
We further compare injection and no-injection designs in all three configurations to separate the effect of learning intermediate features from that of supplying auxiliary predictions as additional action conditions.

Across LIBERO and RoboCasa, all GIFT configurations preserve competitive in-distribution performance, while LIBERO-Plus evaluates robustness under distribution shifts.
Under zero-shot transfer to LIBERO-Plus~\cite{fei2025libero}, all three instantiations outperform their matched baselines across seven distribution shifts; on RoboCasa~\cite{nasiriany2024robocasa}, the gains are especially pronounced for the WAM variants on articulated-object tasks.
Real-robot evaluations further demonstrate robustness under unseen visual and spatial perturbations.
Single-guidance ablations first establish the complementary effects of geometry, affordance, and goal supervision; injection ablations then show that the no-injection variants perform better; and attention visualizations further reveal more consistent focus on task-relevant objects and interaction regions.
Together, these results indicate that GIFT internalizes task-relevant structure into intermediate visual features without relying on auxiliary predictions as additional action conditions.

Our main contributions are:
\begin{itemize}
    \item We identify an action-sufficiency gap shared by VLAs and WAMs and address it by structuring intermediate features through complementary geometry, affordance, and goal supervision.
    \item We instantiate the same objectives as GIFT-VLA, GIFT-WAM-Fast, and GIFT-WAM-IDM, using direct prediction, direct action diffusion, and future-conditioned inverse dynamics, respectively, while keeping action generation model-specific.
    \item Extensive simulation and real-robot evaluations demonstrate consistent gains over matched baselines, with pronounced advantages in zero-shot transfer, articulated-object interaction, and high-precision manipulation.
\end{itemize}

\section{Related Work}

\subsection{Vision-Language-Action Policies and Structured Guidance}

Vision-Language-Action (VLA) policies couple pretrained visual-language representations with robot actions,
extending beyond vision-only controllers~\cite{zhao2023learning,chi2025diffusion}.
Early systems transferred web knowledge to actions~\cite{zitkovich2023rt} and trained foundation policies on large robot datasets~\cite{kim2025openvla}.
Recent models improve continuous control via flow matching and action chunking~\cite{black2024pi0,black2025pi05,kim2025fine}, alongside work on cross-embodiment data~\cite{o2024open,li2024vision} and efficient policies~\cite{pertsch2025fast,shukor2025smolvla}.
However, action-only objectives do not specify which spatial, interaction, and task evidence intermediate features must preserve~\cite{jia2026guidedvla}.

Structured VLA learning addresses this gap with task-relevant guidance.
SpatialVLA~\cite{qu2025spatialvla}, GeoVLA~\cite{sun2025geovla}, and BridgeVLA~\cite{li2025bridgevla} introduce 3D-aware positional or projected representations, Multi-view-VLA~\cite{xiao2026learning} aggregates geometry-guided latent views, and Spatial Forcing~\cite{li2025spatial} aligns visual features with a geometric teacher.
RoboPoint~\cite{yuan2024robopoint} grounds spatial affordances, and CoA-VLA~\cite{li2025coavlaimprovingvisionlanguageactionmodels} combines goal-related object selection with grasp and collision-avoidance cues.
CoT-VLA~\cite{zhao2025cot} externalizes visual reasoning traces, whereas ReconVLA~\cite{song2026reconvla} reconstructs gaze regions to emphasize instruction-relevant targets.
DreamVLA~\cite{zhang2025dreamvla} forecasts compact dynamic, spatial, and semantic knowledge.
GuidedVLA~\cite{jia2026guidedvla} assigns grounding, skill recognition, and geometry perception to specialized decoder heads.

\subsection{World-Action Models and Predictive Representation Learning}

World-Action Models (WAMs) augment action prediction by modeling scene evolution under intervention~\cite{wang2026worldactionmodelsfrontier}.
They connect instructions and actions through future observations or latent states using joint video-action learning~\cite{zhu2025unified,liang2025video}, autoregressive world-action prediction~\cite{cen2025worldvla}, or pretrained video generators~\cite{kim2025cosmospolicy,liao2025genie}.
Joint prediction supplies dynamic supervision and video priors; Fast-WAM~\cite{yuan2026fast} improves features through video co-training without inference-time frame generation.
Nevertheless, future RGB latents optimized for pixel reconstruction can overrepresent appearance while omitting the geometry, object semantics, and task structure needed for action~\cite{su2026worldguidance,yan2026flex}.
Latent-action models compress future motion for video-scale pretraining~\cite{ye2024lapa,routray2025vipra,liang2025clam}.
Yet coarse latent actions may lack detail for precise control~\cite{su2026worldguidance}.

Recent WAMs therefore introduce task-aware prediction targets and conditioning spaces.
GAM~\cite{han2026geometric} grounds temporal prediction and action decoding in geometric foundation features.
GeoSem-WAM~\cite{ma2026geosemwamgeometrysemanticawareworld} jointly predicts future RGB, geometry, and semantic representations; and Flex-$\pi$~\cite{yan2026flex} jointly denoises RGB, 3D pointmaps, and object-level semantics in a shared latent space.
Object-centric methods provide another form of structure: MaskWAM~\cite{yu2026maskwam}
uses target-object masks as prompts and prediction targets,
SG-WAM generates text-grounded object and spatial
foresight~\cite{he2026sgwamtextgroundedspatialawaresemantic}, and OA-WAM represents robots and objects
as instruction-addressable slots~\cite{liu2026oawamobjectaddressableworldaction}.
World Guidance~\cite{su2026worldguidance} compresses future observations into action-conditioning targets, while DIAL~\cite{chen2026dial} predicts latent visual foresight as an intent bottleneck.

\subsection{Distinction between GIFT and Prior Work}

GIFT differs from prior work in how structured knowledge is organized and applied.
\textbf{Decoder specialization.}
GuidedVLA~\cite{jia2026guidedvla} assigns grounding, geometry perception, and skill recognition to specialized VLA decoder heads.
GIFT instead supervises intermediate visual features shared with the action pathway, without fixing head functions or decoder architecture.
\textbf{Future-prediction interfaces.}
DreamVLA~\cite{zhang2025dreamvla} forecasts dynamic, spatial, and semantic knowledge, World Guidance~\cite{su2026worldguidance} learns a compact future-observation condition space, and DIAL~\cite{chen2026dial} uses visual foresight as an inverse-dynamics intent bottleneck.
GIFT treats geometry, affordance, and goal as training constraints on intermediate features; its default no-injection form excludes auxiliary predictions from the action generator.
This separation attributes improvements to representation shaping rather than extra inference-time conditions, while allowing the same objectives to guide policies with different action inputs.
\textbf{Concurrent affordance-centric preprints.}
AffordanceVLA~\cite{yu2026affordancevla} predicts Which2Act, Where2Act, and How2Act cues through specialized experts, while AffordVLA~\cite{kong2026affordvla} aligns VLA features with an affordance teacher.
These concurrent efforts likewise demonstrate the value of shaping intermediate features for control.
GIFT generalizes this idea beyond affordance-specific VLA designs: geometry--affordance--goal jointly encode motion feasibility, interaction selection, and instruction-conditioned relevance across a semantics-centered VLA, a direct-action WAM, and an inverse-dynamics WAM, while preserving their action formulations.

Our preliminary PokeVLA~\cite{zheng2026pokevla} combined geometry alignment and goal-region supervision in a compact VLA.
GIFT adds object-centric affordance supervision, unifies all three signals as a principle for training intermediate features, and validates it across the three policy formulations above.
Matched backbones, a same-head VLA control, and injection/no-injection variants isolate representation supervision from action-head changes and auxiliary conditioning.
Thus, GIFT contributes an architecture-flexible principle with cross-formulation evidence rather than an architecture-specific auxiliary interface.

\section{Methodology}
\label{sec:method}

\begin{figure*}[t]
\centering
\includegraphics[width=\linewidth]{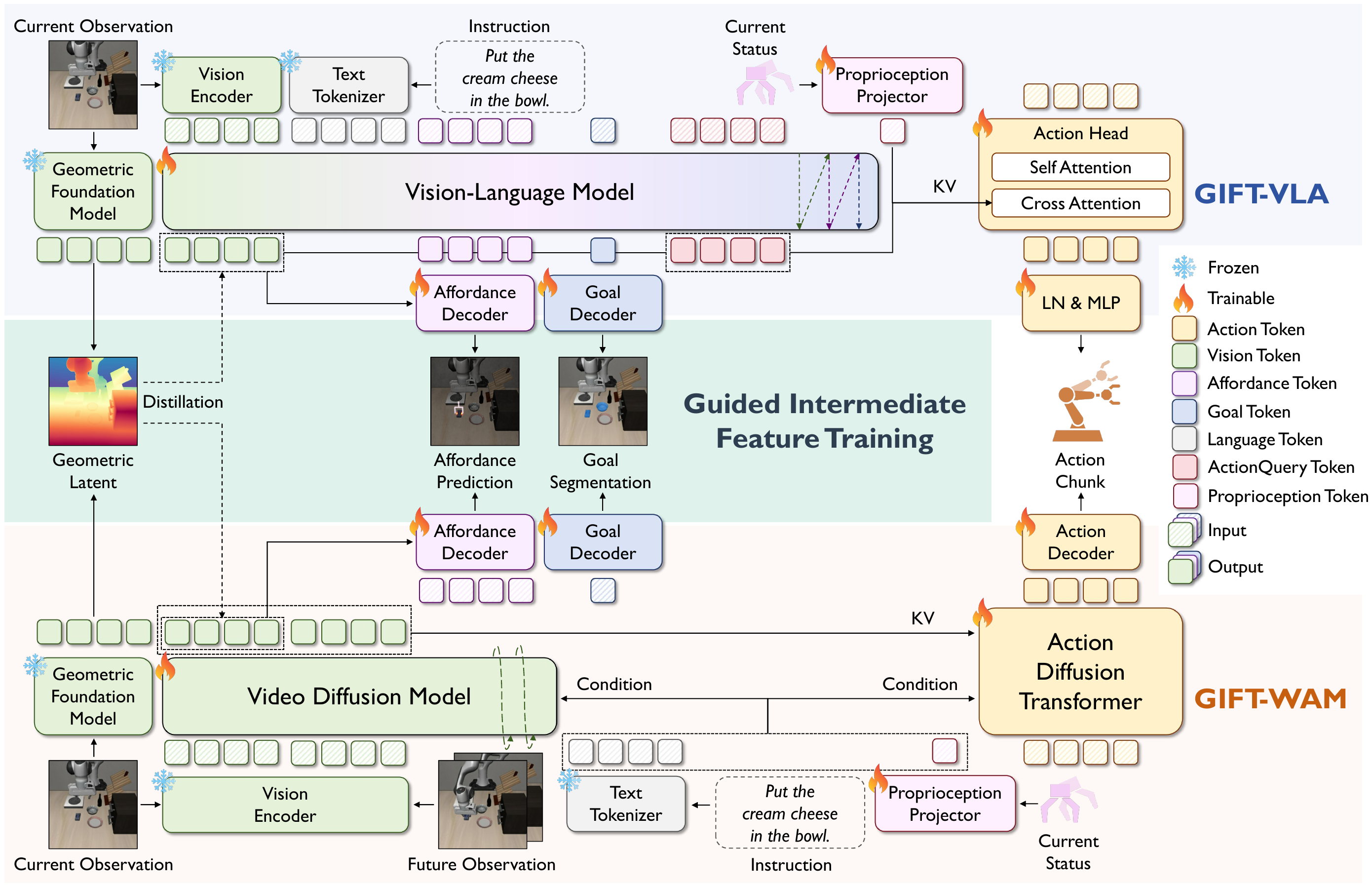}
\caption{\textbf{Overview of GIFT.}
Given the observation, instruction and robot state, GIFT attaches three training objectives to the intermediate visual tokens of a policy: geometry alignment with a frozen geometric foundation model, object-centric affordance prediction, and instruction-conditioned goal-region prediction.
The default no-injection design does not route auxiliary predictions into the action pathway; the auxiliary objectives shape the shared intermediate features only through their training gradients.
We instantiate the same principle in a VLA (top) and two WAM action formulations (bottom).}
\label{fig:main}
\end{figure*}

\subsection{Problem Formulation and Overview}
\label{sec:method_overview}

Consider a robot manipulation sample containing multi-view observations
$O_t=\{I_t^v\}_{v=1}^{V}$, a language instruction $l$, the current proprioceptive state $s_t$, and a future action chunk
$A_t=(a_t,a_{t+1},\ldots,a_{t+H-1})\in\mathbb{R}^{H\times d_a}$.
Here, $t$ is the current time step, $v\in\{1,\ldots,V\}$ indexes the camera view, $V$ is the number of views, $H$ is the action horizon, and $d_a$ is the action dimension.
Let $C_t$ collect the policy's native action conditions, including $(l,s_t)$ and, for the inverse-dynamics model, its future-video context.
A policy predicts $\hat A_t\in\mathbb{R}^{H\times d_a}$ from its intermediate features $Z_t$ as
\begin{equation}
    \hat A_t=\mathcal{A}_{\phi}(Z_t,C_t), \qquad
    Z_t=\mathcal{E}_{\theta}(O_t,l,s_t).
    \label{eq:policy_formulation}
\end{equation}
Here, $\theta$ and $\phi$ denote the learnable parameters of the encoder $\mathcal{E}_{\theta}$ and the action generator $\mathcal{A}_{\phi}$, respectively.
$\psi_{\mathrm{geo}}$, $\psi_{\mathrm{aff}}$, and $\psi_{\mathrm{goal}}$ denote the learnable parameters of the three auxiliary decoders (their subscripts are suppressed on $\mathcal D$ for readability); the VGGT teacher is frozen.
$\mathcal{E}_{\theta}$ can be a vision--language backbone or the video branch of a world--action model, and $\mathcal{A}_{\phi}$ can be a direct regression head, a diffusion action expert, or an inverse-dynamics module.
Despite their rich visual knowledge, these backbones do not necessarily produce control-oriented intermediate features.
As illustrated in Fig.~\ref{fig:main}, GIFT addresses this gap with complementary
\emph{geometry}, \emph{affordance}, and \emph{goal} supervision. These signals
respectively capture motion feasibility, instruction-relevant
object--end-effector relations, and instruction-conditioned task-relevant regions.
Let $r$ denote a selected backbone layer and
$Z_t^{(r)}=\{z_{t,i}^{(r)}\}_{i=1}^{N_r}\in\mathbb{R}^{N_r\times d_r}$ denote its visual-token subset, where $i$ indexes the $N_r$ spatial tokens after flattening the view and patch dimensions and $d_r$ is the token width.
We use shallow features ($r=6$) for geometry guidance and semantically richer final-layer features ($r=-1$) for affordance and goal segmentation.
Lightweight prediction heads $\mathcal{D}_{\mathrm{geo}}$, $\mathcal{D}_{\mathrm{aff}}$, and $\mathcal{D}_{\mathrm{goal}}$ are trained to predict the corresponding structured targets from these selected current-frame features.
The default GIFT formulation is the \emph{no-injection} design:
\begin{equation}
    \hat A_t=\mathcal{A}_{\phi}(Z_t,C_t),
    \qquad
    C_t=C_t^{\mathrm{native}},
    \label{eq:no_injection}
\end{equation}
where $C_t^{\mathrm{native}}$ denotes the policy's original action conditions; the auxiliary geometry, affordance, and goal predictions $\hat G_t$, $\hat U_t$, and $\hat M_t$ are excluded from the action inputs and shape the shared intermediate features only through their backpropagated losses.
This separation lets the same objectives supervise heterogeneous policy families while keeping action generation model-specific.

\subsection{Structured Intermediate Guidance}
\label{sec:structured_guidance}

\subsubsection{Geometry Guidance}
\label{sec:geometry_guidance}

Geometry determines whether a candidate motion is physically feasible, but RGB reconstruction and action supervision alone need not preserve the spatial structure encoded by a geometric model.
We therefore use a frozen VGGT geometric foundation model~\cite{wang2025vggt} as a training-time teacher.
For the same set of input views, the teacher produces patch features
$G_t=\{g_{t,i}\}_{i=1}^{N_g}$, where $g_{t,i}\in\mathbb R^{d_g}$, $d_g$ is the VGGT feature width, $i$ indexes the view--patch locations flattened across all views, and $N_g$ is their total number after resampling.
We spatially resample the teacher grid to establish a one-to-one correspondence with the selected student-token grid, so that $N_g=N_r$.
Because directly regressing features from two heterogeneous backbones can be sensitive to their different feature scales, we decompose each teacher feature into its direction and magnitude.
A layer-normalized two-layer MLP and a scalar scale head predict these two components from each student token:
\begin{equation}
    (\hat g_{t,i},\hat \rho_{t,i})
    =\mathcal{D}_{\mathrm{geo}}(z_{t,i}^{(r)}),
    \qquad i=1,\ldots,N_g.
    \label{eq:geometry_decoder}
\end{equation}
Here, $\hat g_{t,i}\in\mathbb R^{d_g}$ is the projected student feature and $\hat\rho_{t,i}\in\mathbb R$ is its predicted log-scale; we collect the former as $\hat G_t=\{\hat g_{t,i}\}_{i=1}^{N_g}$.
Define the unit directions and compressed teacher magnitude as
\begin{align}
    d_{t,i}&=\frac{g_{t,i}}{\max(\|g_{t,i}\|_2,\epsilon)},\qquad
    \hat d_{t,i}=\frac{\hat g_{t,i}}{\max(\|\hat g_{t,i}\|_2,\epsilon)},\nonumber\\
    \rho_{t,i}&=\log(1+\|g_{t,i}\|_2),
    \label{eq:geometry_components}
\end{align}
where $d_{t,i},\hat d_{t,i}\in\mathbb R^{d_g}$, $\rho_{t,i}\in\mathbb R$, and $\epsilon>0$ ensures numerical stability.
The angular alignment loss is
\begin{equation}
    \mathcal{L}_{\mathrm{ang}}
    =\frac{1}{N_g}\sum_{i=1}^{N_g}
    \left(1-\hat d_{t,i}^{\top}d_{t,i}\right),
    \label{eq:geometry_angular_loss}
\end{equation}
and the scale alignment loss is
\begin{equation}
    \mathcal{L}_{\mathrm{scale}}
    =\frac{1}{N_g}\sum_{i=1}^{N_g}
    \left(\hat\rho_{t,i}-\rho_{t,i}\right)^2.
    \label{eq:geometry_scale_loss}
\end{equation}
The former is invariant to feature norm and aligns the orientation of the student prediction with the VGGT feature, whereas the latter retains the magnitude information discarded by cosine alignment.
Their weighted sum gives the geometry objective
\begin{equation}
    \mathcal{L}_{\mathrm{geo}}
    =\lambda_{\mathrm{ang}}\mathcal{L}_{\mathrm{ang}}
    +\lambda_{\mathrm{scale}}\mathcal{L}_{\mathrm{scale}},
    \label{eq:geometry_loss}
\end{equation}
where we set the corresponding loss weights to
$\lambda_{\mathrm{ang}}=0.2$ and $\lambda_{\mathrm{scale}}=0.05$ in all experiments.
Together, the two predictions define the scale-reconstructed feature
\begin{equation}
    \tilde g_{t,i}
    =\hat d_{t,i}\left(\exp(\hat\rho_{t,i})-1\right).
    \label{eq:geometry_reconstruction}
\end{equation}
We examine the teacher compatibility of $\tilde g_{t,i}$ through the depth-based diagnostic in Sec.~\ref{sec:analysis}.
Unlike approaches that introduce depth or point clouds as a new inference modality, the geometric teacher is used only to construct the training target.
The aligned visual tokens remain in the policy's original representation space and incur no teacher-model cost at deployment.

\subsubsection{Affordance Guidance}
\label{sec:affordance_guidance}

Geometry alone describes what occupies the scene, but not how the robot can interact with it.
Inspired by HumanEgo's object-centric representation of hand--object
interactions~\cite{wang2026humanego}, we define interaction affordance as an
instruction-conditioned, structured
object--end-effector relation: entity roles identify instruction-relevant objects, while
object-centric end-effector poses and end-effector closure states describe the current
interaction configuration. This representation directly encodes what to interact
with and how.

Specifically, we identify all objects referred to by the language instruction and
choose the first object interacted with in the demonstration as the anchor $a$.
For entity slot $k\in\{1,\ldots,K\}$, we define the 20-D interaction target
\begin{equation}
    u_{t,k}=\left[c_{t,k},\xi^{k\mid a}_{t,k},
    \xi^{e(k)\mid k}_{t,k},f_{t,k}\right]\in\mathbb R^{20}.
    \label{eq:affordance_token}
\end{equation}
Here, $c_{t,k}\in\mathbb R$ is a scalar role ID distinguishing padding,
the left or right end effector, the anchor object, and other instruction-relevant
objects. Each pose $\xi=[p,r]\in\mathbb R^9$ comprises a 3-D translation
$p$ and a continuous 6-D rotation representation $r$.
$\xi^{k\mid a}_{t,k}$ gives entity $k$ in the anchor frame, while
$\xi^{e(k)\mid k}_{t,k}$ gives its associated end effector $e(k)$ in the
entity frame. For bimanual tasks, the language instruction explicitly assigns
the manipulation performed by the left and right arms. We accordingly associate
each instruction-relevant object with its designated arm and express the pose of
that arm's end effector in the object coordinate frame. In the real-robot
experiments, we set $K=4$ for the single-arm xArm7 Tasks~1--2 and $K=5$ for
the dual-arm ARX X5 Tasks~3--4. All allocated slots are valid and supervised,
so no padding or slot-validity masking is used. In the bimanual test-tube task,
the five slots correspond to the two end effectors and three
instruction-relevant objects.
For an object slot, $f_{t,k}=-1$ denotes a non-applicable closure state.
For an end-effector slot, $\xi^{k\mid a}_{t,k}$ records its pose in the anchor
frame, $\xi^{e(k)\mid k}_{t,k}$ is the identity pose, and $f_{t,k}$ records its
closure state. We use a binary open/closed value for a parallel gripper and the
closure state derived from the distance between the thumb and index fingertips
for a dexterous hand.
For datasets with variable entity counts, we define a slot-validity mask
$q_{t,k}\in\{0,1\}$; unused slots are zero-padded and excluded by $q_{t,k}$.
We assume each sample contains at least one valid slot, so $\sum_kq_{t,k}>0$.
The affordance targets supervise only the
intermediate features; they are neither policy inputs nor required at
inference.
An affordance decoder summarizes the shared visual tokens with $K$ learned queries
and regresses the 20-D interaction targets:
\begin{equation}
    \hat U_t=\mathcal{D}_{\mathrm{aff}}(Z_t^{(r)}).
    \label{eq:affordance_prediction}
\end{equation}
\begin{equation}
    \begin{gathered}
        U_t=\{u_{t,k}\}_{k=1}^{K},\\
        \hat U_t=\{\hat u_{t,k}\in\mathbb R^{20}\}_{k=1}^{K}.
    \end{gathered}
    \label{eq:affordance_decoder}
\end{equation}
Let $Q_t=\sum_{k=1}^{K}q_{t,k}$ and define the element-averaged Smooth L1
distance with $\beta=1$ as
\begin{equation}
    \begin{aligned}
        \operatorname{SL1}(\hat u,u)
        &=\frac{1}{20}\sum_{j=1}^{20}\rho(\hat u_j-u_j),\\
        \rho(d)
        &=\begin{cases}
            \frac{1}{2}d^2, & |d|<1,\\
            |d|-\frac{1}{2}, & \text{otherwise}.
        \end{cases}
    \end{aligned}
    \label{eq:smooth_l1}
\end{equation}
The affordance objective is
\begin{equation}
    \mathcal{L}_{\mathrm{aff}}
    =\frac{1}{Q_t}
    \sum_{k=1}^{K}q_{t,k}\,
    \operatorname{SL1}(\hat u_{t,k},u_{t,k}).
    \label{eq:affordance_loss}
\end{equation}
The anchor-relative and object-centric parameterization explicitly couples
instruction-relevant entities with the end-effector poses required to interact
with them while reducing dependence on the global camera frame.

\subsubsection{Goal Guidance}
\label{sec:goal_guidance}

Affordance describes the structured interaction configurations of
instruction-relevant entities but does not provide dense image-space localization.
Goal guidance supplies this complementary spatial grounding by identifying
task-relevant action regions.
For every input view, we construct a binary target mask from the instruction-relevant manipulated object or region and collect the masks as
$M_t=\{M_t^v\}_{v=1}^{V}$, where $M_t^v\in\{0,1\}^{h_m\times w_m}$ and $(h_m,w_m)$ is the decoder-grid resolution.
Importantly, $M_t$ is the sole goal target and supervises only \emph{where} the intended interaction should occur.

Because $Z_t^{(r)}$ is conditioned on the instruction, a goal decoder can predict an instruction-specific low-resolution mask
$\hat M_t=\mathcal{D}_{\mathrm{goal}}(Z_t^{(r)},l)=\{\hat M_t^v\}_{v=1}^{V}$, with $\hat M_t^v\in[0,1]^{h_m\times w_m}$.
We use binary cross entropy and soft Dice loss:
\begin{align}
    \mathcal{L}_{\mathrm{goal}}
    =&\lambda_{\mathrm{bce}}\operatorname{BCE}(\hat M_t,M_t)
      +\lambda_{\mathrm{dice}}\left(1-\operatorname{Dice}(\hat M_t,M_t)\right),
    \label{eq:goal_loss}
\end{align}
where both terms are averaged over views and pixels, and Dice is defined as
\begin{equation}
    \operatorname{Dice}(\hat M,M)
    =\frac{2\langle\hat M,M\rangle+\epsilon}
    {\|\hat M\|_1+\|M\|_1+\epsilon}.
    \label{eq:dice}
\end{equation}
Here, $\lambda_{\mathrm{bce}},\lambda_{\mathrm{dice}}\geq0$ are loss weights.
The target mask is downsampled to match the decoder grid before computing the loss.
Together, affordance and goal separate two complementary questions: how the
instruction-relevant entities should be interacted with and where the intended
interaction occurs in the image.

\subsection{Instantiation Across Policy Paradigms}
\label{sec:policy_instantiations}

The three objectives above act on intermediate visual tokens and therefore do not assume a particular action parameterization.
We instantiate GIFT on three policies with deliberately different action-generation mechanisms.

\subsubsection{GIFT-VLA: Semantics-Centered Direct Action Policy}
\label{sec:gift_vla}

GIFT-VLA builds on the StarVLA-OFT formulation~\cite{community2026starvla} with a Qwen3-VL backbone.
Multi-view images and the instruction are processed by the VLM, while proprioception is projected as an additional action condition.
We append one goal token, $K$ affordance tokens, and $N_q$ action-query tokens to the multimodal sequence, where $N_q=H$, so that each predicted action step has one query.
Geometry guidance is applied directly to the selected shallow-layer visual tokens.
Each affordance-token feature is decoded by a layer-normalized MLP into the
20-D interaction target in Eq.~\eqref{eq:affordance_token}.
The goal-token feature cross-attends to current visual tokens and is matched against projected visual features to produce $\hat M_t$.
To support the optional injection analysis in Sec.~\ref{sec:aux_injection}, we replace the original MLP regression head with a layer-aligned BridgeAttention action head.
This modification is not required by the GIFT objectives; it provides an interface for exposing auxiliary features to the action generator in the injection variant and is evaluated separately with the same head but without guidance.
As shown in Fig.~\ref{fig:main}(top), the head draws on layer-aligned VLM hidden states and action-query features through separate cross-attention streams.

The action head maintains $N_q$ learned action latents and refines them through successive bridge blocks, with projected proprioception appended to the conditioning stream.
After the final block, the latents are mapped to a continuous action chunk and trained by direct $\ell_1$ regression:
\begin{equation}
    \mathcal{L}_{\mathrm{VLA}}
    =\frac{1}{Hd_a}\sum_{h=0}^{H-1}
    \|\hat a_{t+h}-a_{t+h}\|_1.
    \label{eq:vla_loss}
\end{equation}
In the no-injection model, affordance and goal tokens are explicitly masked from the action head's raw-token cross-attention.
Consequently, the action head cannot consume their predictions directly, and any improvement must arise from the shared VLM features shaped by Eqs.~\eqref{eq:geometry_loss}--\eqref{eq:goal_loss}.

\subsubsection{GIFT-WAM-Fast: Direct Action from World-Model Features}
\label{sec:gift_wam_fast}

GIFT-WAM-Fast follows Fast-WAM~\cite{yuan2026fast}, which couples a pretrained video diffusion transformer with an action diffusion transformer in a Mixture-of-Transformers architecture.
During training, the video expert denoises future visual latents while the action expert denoises an action chunk.
The fast-mode attention pattern allows action tokens to attend to the current-frame video tokens, but not the noisy future-video tokens.
Video prediction nevertheless trains the shared video expert to encode scene dynamics.
At inference, current-frame video features are computed once and cached for iterative action denoising, so the model retains its native direct-action behavior without generating a future video.

We attach all three GIFT objectives to the current-frame portion of the video stream.
Geometry guidance projects the selected shallow-layer video tokens to the VGGT space.
At the final layer, a learned-query affordance decoder pools $K$ interaction features, while a goal decoder maps spatial first-frame tokens to the instruction-conditioned target mask.
The native fast-mode training objective remains unchanged:
for modality $x\in\{\mathrm{vid},\mathrm{act}\}$, let
\begin{equation}
    \mathcal{L}_{\mathrm{FM}}^{x}
    =\mathbb{E}_{\tau,x_{\tau},\chi_x}
    \left[\left\|
    \mathcal{F}^{x}(x_{\tau},\tau;\chi_x)-u^{x}_{\tau}
    \right\|_2^2\right],
    \label{eq:flow_matching}
\end{equation}
where $\tau\in[0,1]$ is the flow time, $x_\tau$ is the corresponding noisy video or action variable, $\chi_x$ denotes its native conditioning context, $\mathcal F^x$ is the predicted vector field, and $u^x_\tau$ is the target conditional flow.
The Fast objective is
\begin{equation}
    \mathcal{L}_{\mathrm{Fast}}
    =\lambda_v\mathcal{L}_{\mathrm{FM}}^{\mathrm{vid}}
    +\lambda_a\mathcal{L}_{\mathrm{FM}}^{\mathrm{act}},
    \label{eq:fast_loss}
\end{equation}
where $\lambda_v,\lambda_a\geq0$ weight the video and action flow-matching objectives.

\subsubsection{GIFT-WAM-IDM: Future-Conditioned Inverse Dynamics}
\label{sec:gift_wam_idm}

The inverse-dynamics variant retains the same video and action experts but changes the information available to action generation.
During training, a noisy branch learns the future-video objective, while a separate teacher-forcing branch provides a clean or corrupted future-video trajectory to the action expert.
The action tokens attend to this conditional video branch and learn the inverse mapping from the predicted transition to the demonstrated action.
At inference, the model first denoises a future visual trajectory and then freezes its video features as key--value context for action denoising.
This preserves the native imagine-then-act mechanism of the IDM baseline.

GIFT supervises the current-frame tokens at the same respective layers as in the fast-mode model; neither the auxiliary targets nor the auxiliary heads replace the future-conditioned action pathway.
Its native objective is again the sum of video and action flow-matching losses, but the action term is conditioned on the future-video branch:
\begin{equation}
    \mathcal{L}_{\mathrm{IDM}}
    =\lambda_v\mathcal{L}_{\mathrm{FM}}^{\mathrm{vid}}
    +\lambda_a\mathcal{L}_{\mathrm{FM}}^{\mathrm{act}\mid\mathrm{vid}}.
    \label{eq:idm_loss}
\end{equation}
Here, $\mathcal{L}_{\mathrm{FM}}^{\mathrm{act}\mid\mathrm{vid}}$ is Eq.~\eqref{eq:flow_matching} with $x=\mathrm{act}$ and $\chi_x$ additionally containing the teacher-forced future-video features during training.
Comparing GIFT-WAM-Fast and GIFT-WAM-IDM therefore tests whether the same representation principle transfers between direct action diffusion and future-conditioned inverse dynamics.

\subsection{Optional Auxiliary-Feature Injection}
\label{sec:aux_injection}

\begin{figure}[!b]
\centering
\includegraphics[width=\linewidth]{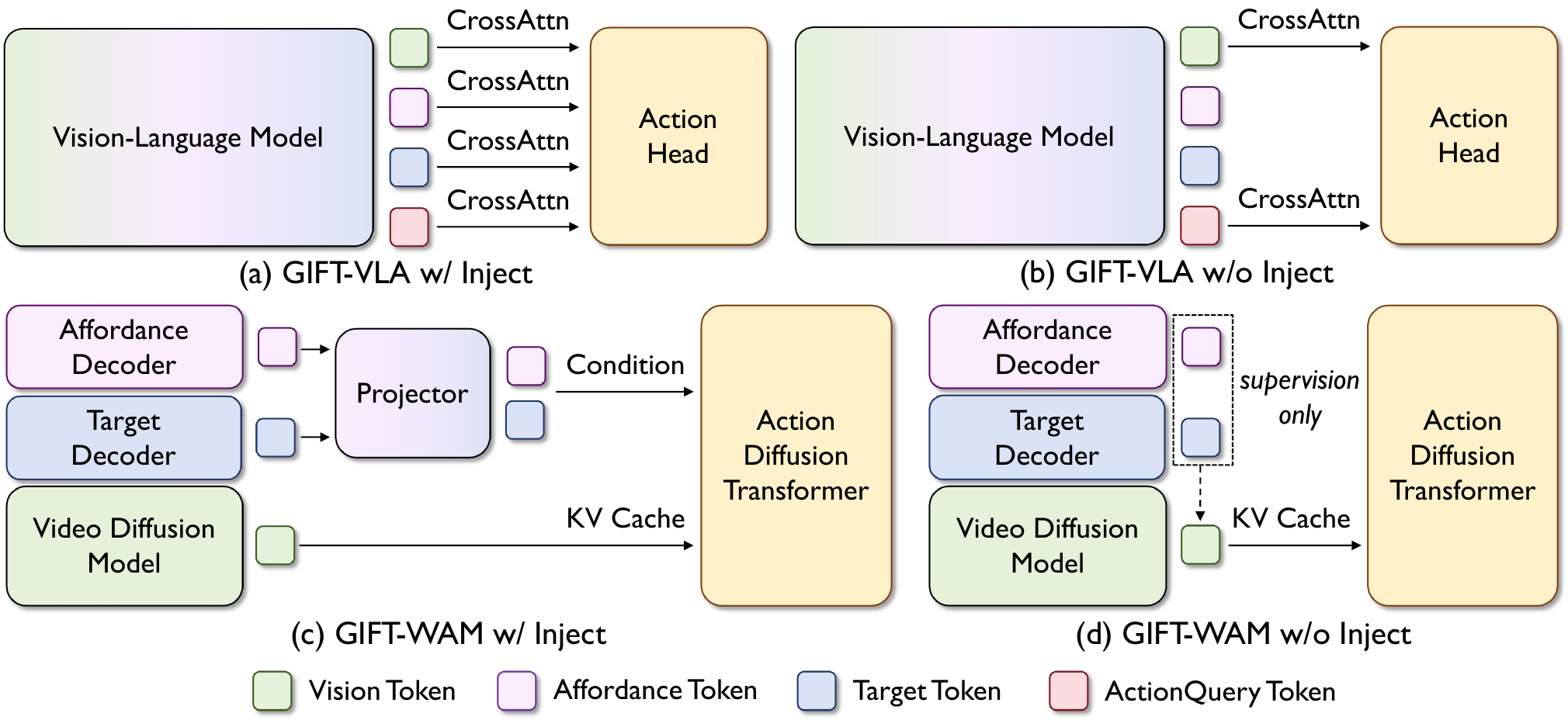}
\caption{\textbf{Injection and no-injection variants.}
With injection, model-predicted affordance and goal features are routed to the model-specific action head during both training and inference.
Without injection, the same predictions provide auxiliary supervision only.
Geometry remains an alignment-only objective because its scene-level structure is absorbed directly by the visual tokens already entering the model-specific action pathway, whereas the object-level affordance and goal features provide compact auxiliary conditions.}
\label{fig:inject}
\end{figure}

The no-injection formulation is the main design of our GIFT because it most directly tests whether structured intermediate supervision alone produces control-oriented intermediate features without routing auxiliary predictions into the action-generation interface.
For completeness, we also implement the injection variant in all three policy families, as shown in Fig.~\ref{fig:inject}.
Let $F_t^{\mathrm{aff}}$ and $F_t^{\mathrm{goal}}$ denote the hidden features used by the affordance and goal decoders before their final prediction layers.
Injection augments the model-specific action condition $C_t$ as
\begin{equation}
    \widetilde C_t=
    \left[C_t;
    \mathcal{P}_{\mathrm{aff}}(F_t^{\mathrm{aff}});
    \mathcal{P}_{\mathrm{goal}}(F_t^{\mathrm{goal}})\right],
    \qquad
    \hat A_t=\mathcal{A}_{\phi}(Z_t,\widetilde C_t),
    \label{eq:injection}
\end{equation}
where $[\,;\,]$ denotes concatenation along the context-token dimension, and the separate projectors $\mathcal{P}_{\mathrm{aff}}$ and $\mathcal{P}_{\mathrm{goal}}$ map their potentially different input widths to the action-conditioning width.
For GIFT-VLA, layer-aligned affordance and goal token features are supplied to the condition cross-attention of each BridgeAttention block.
For both WAM variants, learned-query features from the current-frame video tokens are projected and appended to the action expert's context sequence.

The injection route is active during both training and inference.
At deployment, however, all injected features are predicted internally from the current observation and instruction; no ground-truth auxiliary labels or frozen teacher models are available to the policy.
Geometry is not introduced as an additional conditioning signal, since geometric supervision is applied directly to the visual tokens already consumed through each policy's native visual pathway.
This alignment absorbs geometric information into the native visual stream, allowing the frozen geometry teacher to be discarded at inference and avoiding additional geometry tokens or decoding overhead.
In contrast, affordance and goal describe object-level interaction configurations and task relevance; their decoder features form compact, semantically organized tokens that can be appended as optional action conditions.
This design separates the effect of explicitly routing auxiliary tokens from the representation gains induced by structured intermediate supervision itself.

\subsection{Overall Training Objective}
\label{sec:overall_objective}

For any of the three policy instantiations, the complete objective is
\begin{equation}
    \mathcal{L}_{\mathrm{GIFT}}
    =\mathcal{L}_{\mathrm{native}}
    +\lambda_{\mathrm{geo}}\mathcal{L}_{\mathrm{geo}}
    +\lambda_{\mathrm{aff}}\mathcal{L}_{\mathrm{aff}}
    +\lambda_{\mathrm{goal}}\mathcal{L}_{\mathrm{goal}},
    \label{eq:total_loss}
\end{equation}
where $\mathcal{L}_{\mathrm{native}}$ is selected from
Eqs.~\eqref{eq:vla_loss}, \eqref{eq:fast_loss}, and \eqref{eq:idm_loss}.
The nonnegative scalars $\lambda_{\mathrm{geo}}$, $\lambda_{\mathrm{aff}}$, and $\lambda_{\mathrm{goal}}$ control the three auxiliary objectives, and training optimizes the applicable native parameters $(\theta,\phi)$ together with $(\psi_{\mathrm{geo}},\psi_{\mathrm{aff}},\psi_{\mathrm{goal}})$.
Both injection settings use the same structured targets and overall objective; they differ only in whether the internally predicted affordance and goal features are exposed to the action generator.
In the default no-injection model, the geometry teacher and auxiliary prediction heads do not need to be executed during deployment, and the model-specific action generator receives no auxiliary context.

\section{Simulation Evaluation}

\subsection{Evaluation Protocol}

We evaluate GIFT on three simulation benchmarks that expose complementary aspects
of manipulation performance.

\textbf{LIBERO}~\cite{liu2023libero}
spans four suites, Spatial, Object, Goal, and Long, which respectively
emphasize spatial relations, object-centric knowledge, goal interpretation,
and long-horizon composition. Each suite contains ten language-conditioned
manipulation tasks, performed by a Franka Emika Panda manipulator.
Following prior work, we execute 50 rollouts
per task and report both suite-wise and overall success rates.

\textbf{LIBERO-Plus}~\cite{fei2025libero} expands
LIBERO into seven perturbation distributions defined by shifts in camera viewpoint,
robot initialization, language instruction, lighting, background texture,
sensor noise, and object layout.
We evaluate every perturbed instance once and use the full benchmark evaluation set for both
the main comparison and the ablations.
All models are trained solely on the original
LIBERO training trajectories and receive no fine-tuning
on the augmented LIBERO-Plus training data.

\textbf{RoboCasa}~\cite{nasiriany2024robocasa} GR1 tabletop benchmark
complements LIBERO with a different robot embodiment,
featuring a Fourier GR1
humanoid robot equipped with two 7-DoF arms,
two 6-DoF Fourier dexterous hands, and a 3-DoF waist.
It also provides a visually and dynamically
distinct household-manipulation setting, enabling us to
further assess the effectiveness of GIFT.
The suite contains 24 tasks with 1,000 demonstrations per task: 18 pick-and-place tasks
rearrange objects between supports or receptacles,
whereas six articulated-object tasks additionally require interaction
with an articulated fixture such as a cabinet or a microwave.
We report success over 50 rollouts per task, together with averages for the two
task groups and all tasks.

The GIFT objectives do not change the action-generation interface or provide
auxiliary predictions as action inputs in the default no-injection setting. The
two WAM comparisons therefore use the same backbone and native action-generation
mechanism as their matched baselines and differ only in the
Geometry-Affordance-Goal supervision. For GIFT-VLA, BridgeAttention is introduced
solely to support the optional injection study. Table~\ref{tab:vla-design} provides
a same-head, no-guidance control, separating its 0.7-point architectural effect
from the 3.9-point gain due to structured supervision. The three comparisons are GIFT-VLA versus
StarVLA-OFT, GIFT-WAM-Fast versus Fast-WAM, and GIFT-WAM-IDM versus
Fast-WAM-IDM. Table~\ref{tab:libero} measures standard
in-distribution control, Table~\ref{tab:libero_plus} measures zero-shot transfer
robustness
to challenging environmental perturbations, and Table~\ref{tab:robocasa} further evaluates
performance in a distinct simulation environment and robot embodiment. We report policy success
rate (\%) throughout.

\subsection{Implementation Details}

GIFT is applied during benchmark-specific post-training without an
additional pre-training stage. For GIFT-VLA, we follow the StarVLA-OFT
implementation with a Qwen3-VL-4B-Instruct backbone.
LIBERO and LIBERO-Plus use base and wrist RGB observations, whereas RoboCasa
uses the egocentric view; all images are resized to $224\times224$. GIFT-VLA
predicts 32-step action chunks on all three benchmarks.
We train on 32 GPUs with a per-GPU batch size of 8 (global batch size 256)
for 60,000 optimizer steps.
AdamW uses $\beta=(0.9,0.95)$ and gradient accumulation of 1.
On LIBERO, the base, VLM-interface, and action/auxiliary learning rates are
$2.5\times10^{-5}$, $1\times10^{-5}$, and $1\times10^{-4}$,
with 1,500 warm-up steps and cosine decay to $1\times10^{-6}$. On RoboCasa,
the corresponding rates are $3\times10^{-5}$, $1\times10^{-5}$,
and $1\times10^{-4}$, with 5,000 warm-up steps and cosine decay to $5\times10^{-7}$.

\begin{table}[H]
\caption{Quantitative comparison on LIBERO~\cite{liu2023libero}.
Best and
second-best results are shown in \textbf{bold} and \underline{underline},
respectively.
}
\centering
\renewcommand{\arraystretch}{0.86}
\resizebox{\columnwidth}{!}{%
\begin{tabular}{l c cccc c}
\toprule
\multirow{2}{*}{Method} & 
\multirow{2}{*}{Param. (B)} & \multicolumn{4}{c}{Task Suite} & \multirow{2}{*}{Total} \\
\cmidrule(lr){3-6}
 & & Spatial & Object & Goal & Long & \\
\midrule
OpenVLA~\cite{kim2025openvla} & 7 & 84.7 & 88.4 & 79.2 & 53.7 & 76.5 \\
OpenVLA-OFT~\cite{kim2025fine} & 7 & 97.6 & 98.4 & 97.9 & 94.5 & 97.1 \\
CoT-VLA~\cite{zhao2025cot} & 7 & 87.5 & 91.6 & 87.6 & 69.0 & 81.1 \\
UniVLA~\cite{bu2025univla} & 7 & 96.5 & 96.8 & 95.6 & 92.0 & 95.2 \\
WorldVLA~\cite{cen2025worldvla} & 7 & 87.6 & 85.2 & 75.1 & 54.1 & 74.8 \\
Spatial Forcing~\cite{li2025spatial} & 7 & \underline{99.4} & 99.6 & 98.8 & 96.0 & 98.5 \\
Multi-view-VLA~\cite{xiao2026learning} & 4 & 98.8 & \underline{99.8} & \underline{99.0} & 96.6 & \underline{98.6} \\
4D-VLA~\cite{zhang20254d} & 4 & 88.9 & 95.2 & 90.9 & 79.1 & 88.6 \\
SpatialVLA~\cite{qu2025spatialvla} & 4 & 88.2 & 95.2 & 90.9 & 79.1 & 88.6 \\
StarVLA-OFT~\cite{community2026starvla} & 4 & 97.8 & 98.6 & 96.2 & 93.8 & 96.6 \\
StarVLA-$\alpha$~\cite{ye2026starvla} & 4 & 99.0 & \underline{99.8} & 98.5 & 94.1 & \textbf{98.8} \\
$\pi_0$~\cite{black2024pi0} & 3 & 96.8 & 98.8 & 95.8 & 85.2 & 94.2 \\
$\pi_0$-FAST~\cite{pertsch2025fast} & 3 & 96.4 & 96.8 & 88.6 & 60.2 & 85.5 \\
ACoT-VLA~\cite{zhong2026acot} & 3 & 98.6 & 99.0 & \textbf{99.4} & \underline{97.0} & \underline{98.5} \\
SmolVLA~\cite{shukor2025smolvla} & 2 & 93.0 & 94.0 & 91.0 & 77.0 & 88.8 \\
GR00T N1~\cite{bjorck2025gr00t} & 2 & 94.4 & 97.6 & 93.0 & 90.6 & 93.9 \\
VLA-Adapter~\cite{wang2025vlaadapter} & 1 & \textbf{99.6} & 99.6 & 98.2 & 96.4 & 98.5 \\
DreamVLA~\cite{zhang2025dreamvla} & 0.5 & 97.5 & 94.0 & 89.5 & 89.5 & 92.6 \\
\midrule
Fast-WAM~\cite{yuan2026fast} & 6 & 98.2 & \textbf{100.0} & 97.0 & 95.2 & 97.6 \\
Fast-WAM-IDM~\cite{yuan2026fast} & 6 & 98.8 & 97.8 & 97.8 & \textbf{97.6} & 98.0 \\
Cosmos Policy~\cite{kim2025cosmospolicy} & 2 & 98.1 & \textbf{100.0} & 98.2 & \textbf{97.6} & 98.5 \\
\midrule
\textbf{GIFT-VLA} & 4 & 99.0 & 99.2 & 98.4 & 94.8 & 97.9 \\
\textbf{GIFT-WAM-Fast} & 6 & 97.8 & 99.6 & 98.6 & 94.8 & 97.7 \\
\textbf{GIFT-WAM-IDM} & 6 & 99.0 & \textbf{100.0} & 98.6 & 96.4 & 98.5 \\

\bottomrule
\end{tabular}%
}
\label{tab:libero}
\end{table}

\begin{table*}[!b]
\caption{Quantitative comparison on LIBERO-Plus~\cite{fei2025libero} in a zero-shot transfer setting. All methods are trained only on the standard LIBERO
dataset without fine-tuning on the LIBERO-Plus dataset.}
\centering
\renewcommand{\arraystretch}{0.86}
\setlength{\tabcolsep}{0.018\linewidth}
\begin{tabular}{l c ccccccc c}
\toprule
\multirow{2}{*}{Method} &
\multirow{2}{*}{Param. (B)} & \multicolumn{7}{c}{Perturbation Type} & \multirow{2}{*}{Total} \\
\cmidrule(lr){3-9}
 & & Camera & Robot & Language & Light & Background & Noise & Layout & \\
\midrule
OpenVLA~\cite{kim2025openvla} & 7 & 0.8 & 3.5 & 23.0 & 8.1 & 34.8 & 15.2 & 28.5 & 15.6 \\
OpenVLA-OFT~\cite{kim2025fine} & 7 & 56.4 & 31.9 & 79.5 & 88.7 & 93.3 & 75.8 & 74.2 & 69.6 \\
UniVLA~\cite{bu2025univla} & 7 & 1.8 & 46.2 & 69.6 & 69.0 & 81.0 & 21.2 & 31.9 & 42.9 \\
WorldVLA~\cite{cen2025worldvla} & 7 & 0.1 & 27.9 & 41.6 & 43.7 & 17.1 & 10.9 & 38.0 & 25.0 \\
RIPT-VLA~\cite{tan2025interactive} & 7 & 55.2 & 31.2 & 77.6 & 88.4 & 91.6 & 73.5 & 74.2 & 68.4 \\
Spatial Forcing~\cite{li2025spatial} & 7 & 20.1 & 13.4 & 40.9 & 29.1 & 33.4 & 25.7 & 39.3 & 29.1 \\
Multi-view-VLA~\cite{xiao2026learning} & 4 & \textbf{89.6} & 60.1 & 86.9 & 98.0 & 95.7 & \textbf{97.2} & 78.2 & 85.7 \\
StarVLA-OFT~\cite{community2026starvla} & 4 & 47.0 & 60.1 & 87.0 & 96.3 & 95.3 & 73.1 & 79.2 & 75.0 \\
StarVLA-$\alpha$~\cite{ye2026starvla} & 4 & 48.7 & 63.4 & 86.8 & 95.8 & 94.6 & 75.0 & 80.2 & 77.8 \\
ABot-M0~\cite{yang2026abot} & 4 & 60.4 & 67.9 & 86.4 & 96.2 & 91.6 & 86.4 & 82.6 & 80.5 \\
NORA~\cite{hung2025nora} & 3 & 2.2 & 37.0 & 65.1 & 45.7 & 58.6 & 12.8 & 62.1 & 39.0 \\
$\pi_0$~\cite{black2024pi0} & 3 & 13.8 & 6.0 & 58.8 & 85.0 & 81.4 & 79.0 & 68.8 & 53.6 \\
$\pi_0$-FAST~\cite{pertsch2025fast} & 3 & 65.1 & 21.6 & 61.0 & 73.2 & 73.2 & 74.4 & 68.8 & 61.6 \\
GuidedVLA~\cite{jia2026guidedvla} & 3 & 73.7 & 51.4 & 62.6 & 94.6 & 89.0 & 85.2 & 79.9 & 75.4 \\
ACoT-VLA~\cite{zhong2026acot} & 3 & 72.6 & 82.6 & 87.5 & 97.7 & \underline{96.5} & 87.8 & \textbf{88.1} & \underline{86.6} \\
VLA-JEPA~\cite{sun2026vla} & 2 & 63.3 & 67.1 & 85.4 & 95.6 & 93.6 & 66.3 & 85.1 & 79.5 \\
VLA-Adapter~\cite{wang2025vlaadapter} & 1 & 36.2 & 37.9 & 74.6 & 70.6 & 76.1 & 58.0 & 69.7 & 59.1 \\
\midrule
Fast-WAM~\cite{yuan2026fast} & 6 & 24.9 & 50.7 & 76.9 & 89.2 & 62.3 & 58.0 & 67.7 & 60.0 \\
Fast-WAM-IDM~\cite{yuan2026fast} & 6 & 65.4 & \underline{86.0} & 89.5 & 96.3 & 74.4 & 87.6 & 80.7 & 82.6 \\
ABot-M0.5~\cite{chen2026abot} & 6 & 70.5 & \textbf{87.4} & 88.6 & 94.0 & 89.7 & 75.5 & \underline{85.2} & 83.4 \\
ImageWAM~\cite{zhang2026imagewam} & 4 & 80.8 & 50.3 & \textbf{91.4} & 98.1 & 85.5 & 93.8 & 80.5 & 83.1 \\
Being-H0.7~\cite{luo2026being} & 3 & - & - & - & - & - & - & - & 82.1 \\
Cosmos Policy~\cite{kim2025cosmospolicy} & 2 & 75.8 & 63.3 & 81.7 & 96.5 & 88.9 & 92.7 & 82.2 & 82.2 \\
GAM~\cite{han2026geometric} & 1.4 & \underline{83.1} & 70.0 & 84.8 & 97.2 & 94.3 & \underline{95.3} & 79.1 & 85.5 \\
\midrule

\textbf{GIFT-VLA} & 4 & 57.7 & 63.7 & 88.6 & \underline{98.5} & \textbf{96.9} & 82.5 & 80.3 & 79.6 \\

\textbf{GIFT-WAM-Fast} & 6 & 37.4 & 75.3 & 79.5 & 94.5 & 70.3 & 78.7 & 78.8 & 72.6 \\

\textbf{GIFT-WAM-IDM} & 6 & 78.7 & 80.5 & \underline{90.2} & \textbf{99.0} & 89.2 & \textbf{97.2} & 83.3 & \textbf{87.8} \\
\bottomrule
\end{tabular}
\label{tab:libero_plus}
\end{table*}

For GIFT-WAM, the Fast and IDM variants retain their original action-generation
objectives and pathways: the former directly denoises actions from current-frame
world-model features, whereas the latter conditions inverse dynamics on a predicted
future trajectory.
Empirically, we use 2 denoising steps
at inference: the Fast variant applies them only to the action chunk, whereas
the IDM variant first uses 2 steps to denoise the future video and then 2
steps to denoise the action. Both predict 32-step action chunks and use AdamW
with $\beta=(0.9,0.95)$, a learning rate of $1\times10^{-4}$, weight decay
of $1\times10^{-2}$, a cosine schedule with 5\% warm-up,
and gradient accumulation of 1. On LIBERO, both variants use two $224\times224$
RGB views, 64 GPUs, a per-GPU batch size of 8, and 50 training epochs;
On RoboCasa, they use one $224\times224$ egocentric view, 64 GPUs,
a per-GPU batch size of 16, and train for at most 60,000 optimizer steps.
Following the corresponding baseline protocols, both VLA- and WAM-based policies
use receding-horizon execution, replanning every 10 control steps on LIBERO and
LIBERO-Plus and every 12 control steps on RoboCasa.
Structured targets and the frozen geometry teacher are used only for training.
At inference, no labels or teacher models are required; injection variants
route only the policy's internally predicted affordance and goal features to
the action head.

\textbf{Action spaces and ground-truth actions.}
LIBERO uses a 7-D relative end-effector action
$a_t=[\Delta p_t,\Delta\omega_t,g_t]$, comprising a 3-D Cartesian displacement,
a 3-D axis--angle rotation increment in radians, and a scalar gripper command.
RoboCasa executes a 29-D joint target comprising 7 joints for each arm, 6 for
each hand, and 3 for the waist:
$a_t^{\mathrm{ctrl}}=[q_t^{L,a},q_t^{R,a},q_t^{L,h},q_t^{R,h},q_t^w]
\in\mathbb{R}^{7+7+6+6+3}$. Following its pose-augmented data protocol, we use
$\tilde a_t=[a_t^{\mathrm{ctrl}},p_t^L,r_{6D,t}^L,p_t^R,r_{6D,t}^R]
\in\mathbb{R}^{47}$ as the training target, where each appended next-step
end-effector pose contains a 3-D position and the first two columns of its
rotation matrix. These pose channels provide additional supervision, while only
$a_t^{\mathrm{ctrl}}$ is executed. Executable targets are taken from successful
expert demonstrations, the RoboCasa pose channels are recovered through replay,
and padded tail steps are masked from the action loss.

\begin{table}[t]
\caption{Quantitative comparison on RoboCasa~\cite{nasiriany2024robocasa}.}
\centering
\setlength{\tabcolsep}{0.035\linewidth}
\begin{tabular}{l c cc c}
\toprule
Method & Param. (B) & Art. & P\&P & Avg. \\
\midrule
TwinBrainVLA~\cite{yu2026twinbrainvla} & 8 & 56.7 & 53.9 & 54.6 \\
RLDX-1~\cite{kim2026rldx} & 8 & - & - & 58.7 \\
StarVLA-OFT~\cite{community2026starvla} & 4 & 43.7 & 50.4 & 48.8 \\
StarVLA-$\alpha$~\cite{ye2026starvla} & 4 & 49.0 & 55.4 & 53.8 \\
ABot-M0~\cite{yang2026abot} & 4 & 61.7 & 57.1 & 58.3 \\
GR00T N1.6~\cite{bjorck2025gr00t} & 3 & 24.2 & 55.4 & 47.6 \\
DIAL~\cite{chen2026dial} & 3 & 74.3 &  68.9 & 70.2 \\
\midrule
Fast-WAM~\cite{yuan2026fast} & 6 & 62.0 & 78.8 & 74.6 \\
Fast-WAM-IDM~\cite{yuan2026fast} & 6 & 59.7 & 78.7 & 73.9 \\
Being-H0.7~\cite{luo2026being} & 3 & - & - & 49.2 \\
DiT4DiT~\cite{ma2026dit4dit} & 2 & 50.3 & 50.9 & 50.8 \\
LDA-1B~\cite{lyu2026lda} & 1.6 & 56.3 & 55.1 & 55.4 \\
\midrule
\textbf{GIFT-VLA} & 4 & 60.0 & 61.9 & 61.4 \\
\textbf{GIFT-WAM-Fast} & 6 & \underline{83.3} & \textbf{83.7} & \textbf{83.6} \\
\textbf{GIFT-WAM-IDM} & 6 & \textbf{84.3} & \underline{81.7} & \underline{82.3} \\
\bottomrule
\end{tabular}
\label{tab:robocasa}
\end{table}

\textbf{Loss weights.}
Across policy variants and benchmarks, we set
$\lambda_{\mathrm{ang}}=0.2$, $\lambda_{\mathrm{scale}}=0.05$,
$\lambda_{\mathrm{bce}}=\lambda_{\mathrm{dice}}=0.2$,
$\lambda_v=\lambda_a=1.0$, and
$(\lambda_{\mathrm{geo}},\lambda_{\mathrm{aff}},\lambda_{\mathrm{goal}})=(1.0,0.5,1.0)$.
For single-guidance ablations, we set the weight of each removed objective to
zero and leave all remaining weights unchanged.

\textbf{Simulation structured-target construction.}
For both LIBERO and RoboCasa, we replay expert actions in simulation and retain
only successful trajectories. We extract final-layer patch features from a
frozen VGGT teacher and resample them to the policy's visual-token grid as
geometry-alignment targets. Goal targets are obtained by matching
instruction-associated object IDs to simulator instance segmentations, taking
the union for multiple target objects. Affordance targets are generated from
privileged simulator-provided end-effector and object poses according to
Eq.~\eqref{eq:affordance_token}. We use the language instruction to identify all
task-relevant objects and take the first object interacted with along the expert
trajectory as the anchor. In the simulation experiments, we allocate $K=4$ slots
for the active end effector(s) and instruction-relevant objects. Parallel-gripper states are encoded as binary closure
values, whereas dexterous-hand states encode closure from the distance between
the thumb and index fingertips. Invalid
slots are zero-padded and excluded from both attention and supervision.

\subsection{Comparison with State-of-the-Art Methods}

\textbf{Performance on LIBERO.}
Table~\ref{tab:libero} first examines whether the proposed
intermediate supervision preserves the native control capability
of each policy on the LIBERO benchmark.
GIFT-VLA achieves 97.9\% overall, improving its StarVLA-OFT counterpart
by 1.3 points. GIFT-WAM-Fast and GIFT-WAM-IDM
reach 97.7\% and 98.5\%, exceeding
their corresponding Fast-WAM and Fast-WAM-IDM baselines by 0.1 and 0.5 points, respectively.
These modest margins reflect the near-saturated performance of the matched
baselines on standard LIBERO, which leaves limited headroom; GIFT's advantages
become more pronounced under distribution shifts and fine-grained interaction.
Among all entries, GIFT-WAM-IDM ties the second-best
overall result and attains 100.0\% on Object, matching
the best score in that suite. These results show that imposing
structured intermediate objectives does not compromise
in-distribution control across the VLA, fast-mode WAM,
or inverse-dynamics WAM formulations.

\textbf{Zero-shot robustness on LIBERO-Plus.}
We evaluate zero-shot transfer by training all models only on the original
LIBERO trajectories and directly testing them on LIBERO-Plus without adaptation.
LIBERO-Plus provides a more diagnostic comparison by exposing every policy
to the same seven controlled distribution shifts. As shown in
Table~\ref{tab:libero_plus}, GIFT-WAM-IDM achieves the best overall success
rate of 87.8\%, surpassing ACoT-VLA by 1.2 points and GAM by 2.3 points.
GIFT-VLA reaches 79.6\%, exceeding several
competitive VLA methods such as StarVLA-$\alpha$ and GuidedVLA.
Relative to StarVLA-OFT, Fast-WAM, and Fast-WAM-IDM, GIFT-VLA,
GIFT-WAM-Fast, and GIFT-WAM-IDM improve the overall success rate by 4.6,
12.6, and 5.2 points, respectively.
GIFT-WAM-IDM also records the best results under lighting and sensor-noise perturbations,
while GIFT-VLA obtains the highest score under background changes.
Although the leading method varies
across perturbation types, the paired gains across all three policy paradigms
show that the same representation basis improves
robustness without prescribing a particular action decoder.

\textbf{Performance on RoboCasa.}
RoboCasa benchmark complements LIBERO with a different robot embodiment, visual domain,
and interaction dynamics. As shown in Table~\ref{tab:robocasa},
GIFT-WAM-Fast achieves the best average success rate of 83.6\%,
improving Fast-WAM by 9.0 points and exceeding DIAL by 13.4 points. GIFT-WAM-IDM reaches 82.3\%, an 8.4-point gain over Fast-WAM-IDM. The two variants also set the best group-wise results: GIFT-WAM-IDM obtains 84.3\% on articulated interaction, while GIFT-WAM-Fast reaches 83.7\% on pick-and-place. GIFT-VLA achieves 61.4\%, improving StarVLA-OFT by 12.6 points and outperforming TwinBrainVLA, RLDX-1, StarVLA-$\alpha$, and ABot-M0, while DIAL remains the strongest VLA entry at 70.2\%. On articulated tasks, GIFT-WAM-Fast outperforms Fast-WAM by 21.3 points, and GIFT-WAM-IDM outperforms Fast-WAM-IDM by 24.6 points, showing that structured geometry and affordance cues remain useful under richer contact dynamics and a different embodiment rather than only in the LIBERO domain.

\subsection{Ablation Study and In-depth Analysis}
\label{sec:analysis}

In this section, we organize the ablation and qualitative analysis around
five questions that connect the performance gains to the representations learned by GIFT
and examine their limitations.

\begin{table*}[ht]
\caption{Ablation on geometry, affordance, and goal guidance.}
\centering
\setlength{\tabcolsep}{0.018\linewidth}
\begin{tabular}{ccc ccccccc c}
\toprule
\multirow{2}{*}{Geometry} &
\multirow{2}{*}{Affordance} &
\multirow{2}{*}{Goal} &
\multicolumn{7}{c}{Perturbation Type} &
\multirow{2}{*}{Total} \\
\cmidrule(lr){4-10}
& & & Camera & Robot & Language & Light & Background & Noise & Layout & \\
\midrule
\multicolumn{11}{c}{GIFT-VLA} \\
\cmidrule(lr){1-11}
& & & 47.0 & 60.1 & 87.0 & 96.3 & 95.3 & 73.1 & 79.2 & 75.0 \\
$\checkmark$ & & & 57.4 & 63.4 & 87.5 & 96.9 & 96.8 & 83.0 & 81.1 & 77.1 \\
& $\checkmark$ & & 56.2 & 63.9 & 87.9 & 97.0 & 97.7 & 80.8 & 80.5 & 79.0 \\
& & $\checkmark$ & 58.4 & 62.3 & 87.1 & 97.8 & 96.9 & 81.6 & 80.5 & 79.1 \\
$\checkmark$ & $\checkmark$ & $\checkmark$ & 57.7 & 63.7 & 88.6 & 98.5 & 96.9 & 82.5 & 80.3 & 79.6 \\
\midrule

\multicolumn{11}{c}{GIFT-WAM-Fast} \\
\cmidrule(lr){1-11}
& & & 24.9 & 50.7 & 76.9 & 89.2 & 62.3 & 58.0 & 67.7 & 60.0 \\
$\checkmark$ & & & 38.6 & 70.8 & 74.4 & 93.2 & 65.2 & 77.6 & 77.8 & 70.4 \\
& $\checkmark$ & & 35.3 & 72.6 & 74.7 & 93.2 & 72.5 & 77.5 & 75.2 & 70.5 \\
& & $\checkmark$ & 45.2 & 73.9 & 73.3 & 92.4 & 71.0 & 76.6 & 75.6 & 71.7 \\
$\checkmark$ & $\checkmark$ & $\checkmark$ & 37.4 & 75.3 & 79.5 & 94.5 & 70.3 & 78.7 & 78.8 & 72.6 \\
\midrule

\multicolumn{11}{c}{GIFT-WAM-IDM} \\
\cmidrule(lr){1-11}
& & & 65.4 & 86.0 & 89.5 & 96.3 & 74.4 & 87.6 & 80.7 & 82.6 \\
$\checkmark$ & & & 69.8 & 86.8 & 90.3 & 97.9 & 78.2 & 90.4 & 81.7 & 84.8 \\
& $\checkmark$ & & 68.4 & 91.0 & 91.0 & 96.9 & 79.2 & 89.9 & 83.0 & 85.4 \\
& & $\checkmark$ & 72.3 & 84.7 & 92.8 & 98.4 & 86.5 & 92.7 & 84.0 & 86.9 \\
$\checkmark$ & $\checkmark$ & $\checkmark$ & 78.7 & 80.5 & {90.2} & {99.0} & 89.2 & {97.2} & 83.3 & {87.8} \\
\bottomrule
\end{tabular}
\label{tab:ablation_full}
\end{table*}

\begin{table}[t]
\caption{Ablation on the effectiveness of injecting auxiliary predictions into the action head.}
\centering
\small
\setlength{\tabcolsep}{3pt}
\resizebox{\linewidth}{!}{%
\begin{tabular}{l cc cc}
\toprule
\multirow{2}{*}{Method} &
\multicolumn{2}{c}{LIBERO-Plus} &
\multicolumn{2}{c}{RoboCasa} \\
\cmidrule(lr){2-3}\cmidrule(lr){4-5}
& w/ Inject & w/o Inject & w/ Inject & w/o Inject \\
\midrule
\textbf{GIFT-VLA} & 76.7 & 79.6 & 58.6 & 61.4 \\
\textbf{GIFT-WAM-Fast} & 72.3 & 72.6 & 83.4 & 83.6 \\
\textbf{GIFT-WAM-IDM} & 87.6 & 87.8 & 82.0 & 82.3 \\
\bottomrule
\end{tabular}%
}
\label{tab:injection_ablation}
\end{table}

\textbf{Q1: How does each guidance signal contribute to robustness?}
Table~\ref{tab:ablation_full} isolates geometry, affordance,
and goal guidance under the same LIBERO-Plus protocol.
For every GIFT configuration, each individual signal improves the overall success
rate over its named baseline: StarVLA-OFT for GIFT-VLA, Fast-WAM for
GIFT-WAM-Fast, and Fast-WAM-IDM for GIFT-WAM-IDM.
For GIFT-VLA, the gains from geometry, affordance, and goal are 2.1, 4.0,
and 4.1 points; for GIFT-WAM-Fast, they are 10.4, 10.5, and 11.7 points;
and for GIFT-WAM-IDM, they are 2.2, 2.8, and 4.3 points.
The per-perturbation results further reveal distinct emphases.
Geometry guidance consistently helps camera-viewpoint changes, with gains of 10.4,
13.7, and 4.4 points for GIFT-VLA, GIFT-WAM-Fast, and GIFT-WAM-IDM,
respectively. Preserving surfaces, object extent, and spatial relations
therefore reduces dependence on viewpoint-specific appearance and supports motion
feasibility across camera shifts.
Affordance guidance produces large and consistent gains under robot-initialization
shifts, improving success by 3.8, 21.9, and 5.0 points for GIFT-VLA,
GIFT-WAM-Fast, and GIFT-WAM-IDM, respectively. This pattern suggests that capturing
instruction-relevant entity roles and object-centric interaction configurations
helps the policy adapt to unfamiliar robot starting configurations.
Goal guidance provides particularly strong improvements under camera, background,
and sensor-noise perturbations: the respective gains are 11.4/1.6/8.5 points for
GIFT-VLA, 20.3/8.7/18.6 points for GIFT-WAM-Fast, and 6.9/12.1/5.1 points for
GIFT-WAM-IDM. These gains indicate that grounding instructions in task-relevant
action regions helps preserve task relevance despite changes in viewpoint,
background, and sensor observations.
Combining all three signals yields the best overall success rate for every
configuration: 79.6\% for GIFT-VLA, 72.6\% for GIFT-WAM-Fast, and 87.8\% for GIFT-WAM-IDM,
corresponding to gains of 4.6, 12.6, and 5.2 points over StarVLA-OFT,
Fast-WAM, and Fast-WAM-IDM, respectively. Together, their consistent advantage across all three ablation groups shows that these
complementary signals form a more complete control-oriented representation across
policy paradigms.

\begin{figure*}[t]
\centering
\includegraphics[width=\textwidth]{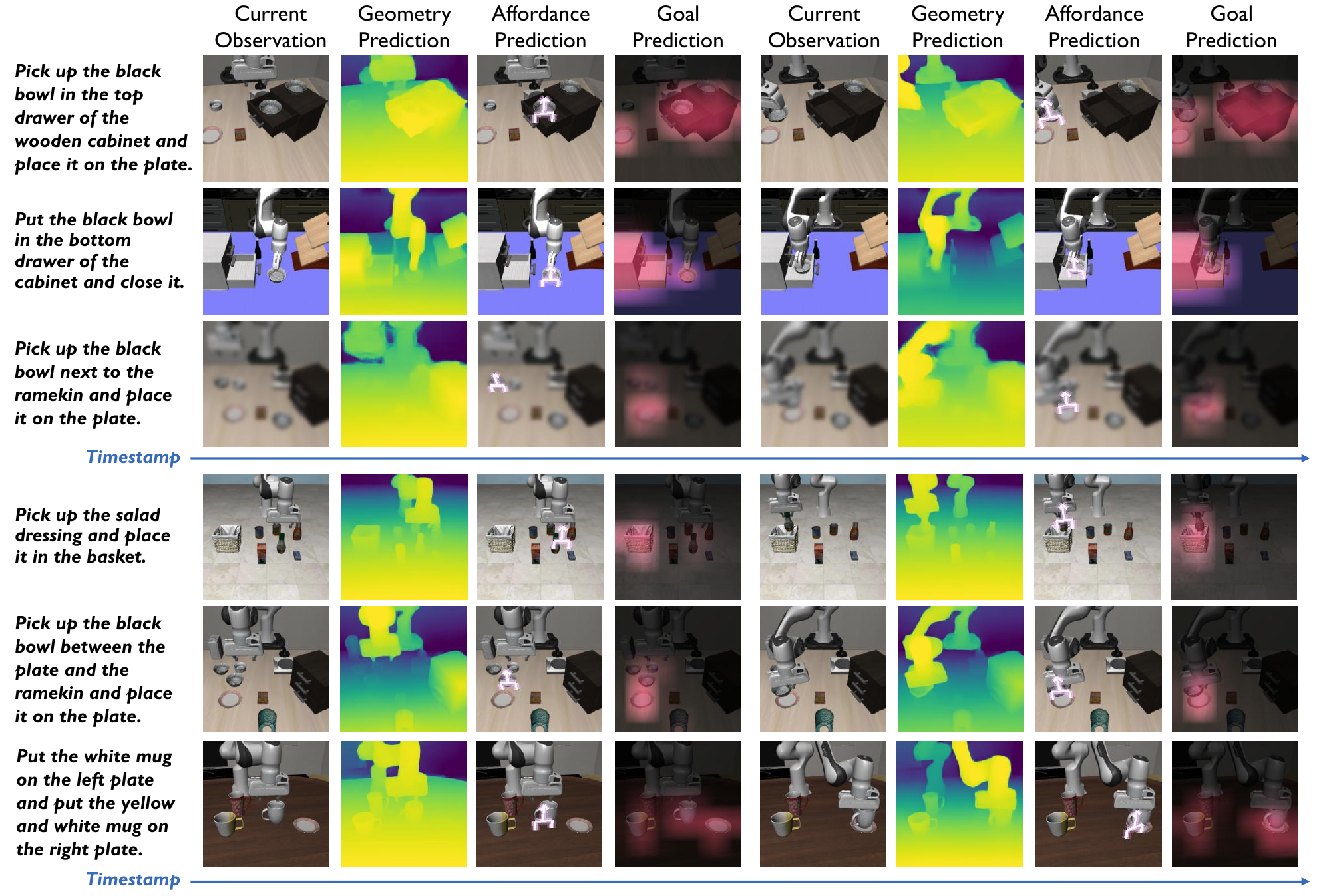}
\caption{\textbf{Visualization of the structured intermediate guidance.} Examples from GIFT-VLA (top) and GIFT-WAM-IDM (bottom) show observations together with geometry, affordance, and goal predictions.}
\label{fig:guidance}
\end{figure*}

\begin{figure*}[t]
\centering
\includegraphics[width=0.96\textwidth]{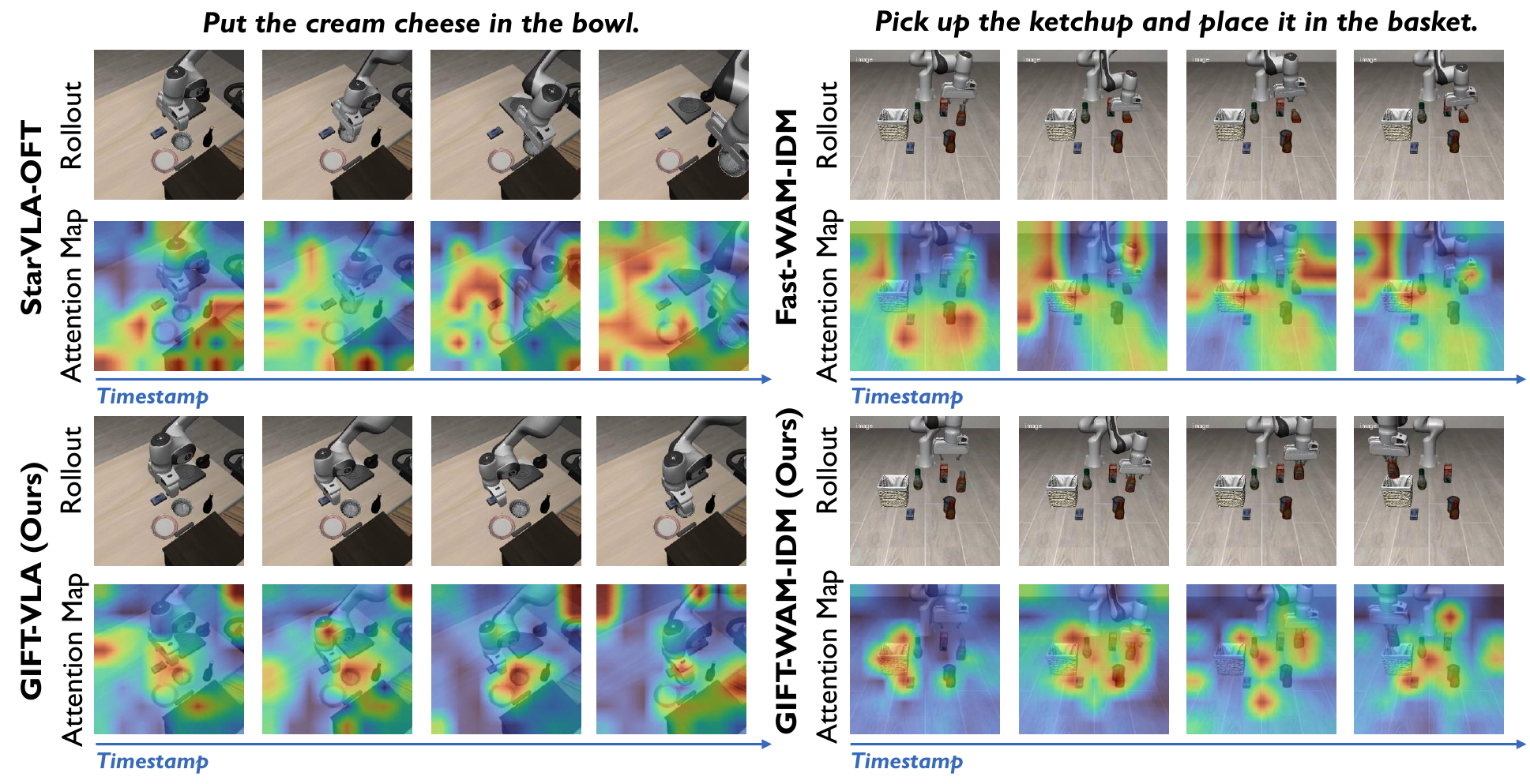}
\caption{\textbf{Visualization of action-to-vision attention maps on LIBERO-Plus.} Across the VLA and WAM-IDM examples, GIFT focuses more consistently on
manipulated objects and interaction-relevant regions compared with matched baselines.}
\label{fig:attention-liberoplus}
\end{figure*}

\begin{figure*}[t]
\centering
\includegraphics[width=0.96\textwidth]{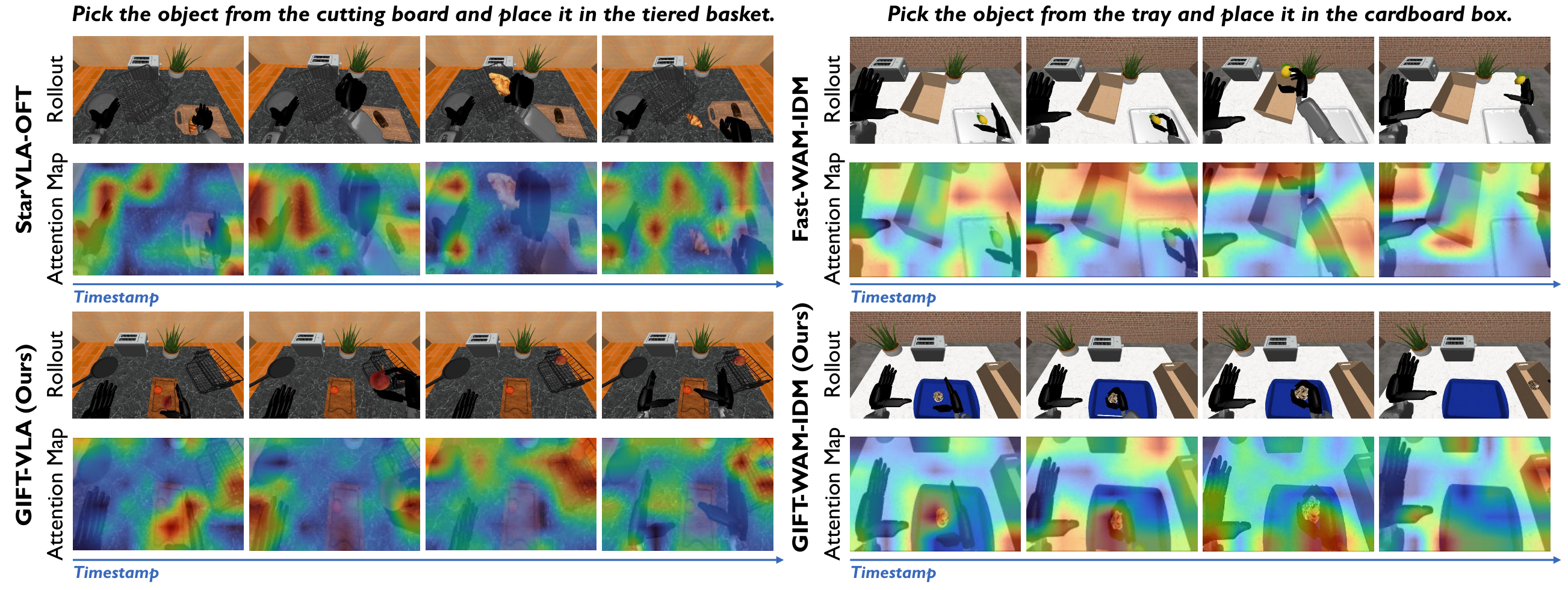}
\caption{\textbf{Visualization of action-to-vision attention maps on RoboCasa.}
Under RoboCasa's distinct humanoid embodiment and household scenes, GIFT still more
consistently concentrates attention on task-relevant regions than matched baselines.}
\label{fig:attention-robocasa}
\end{figure*}

\begin{figure}[t]
\centering
\includegraphics[width=\linewidth]{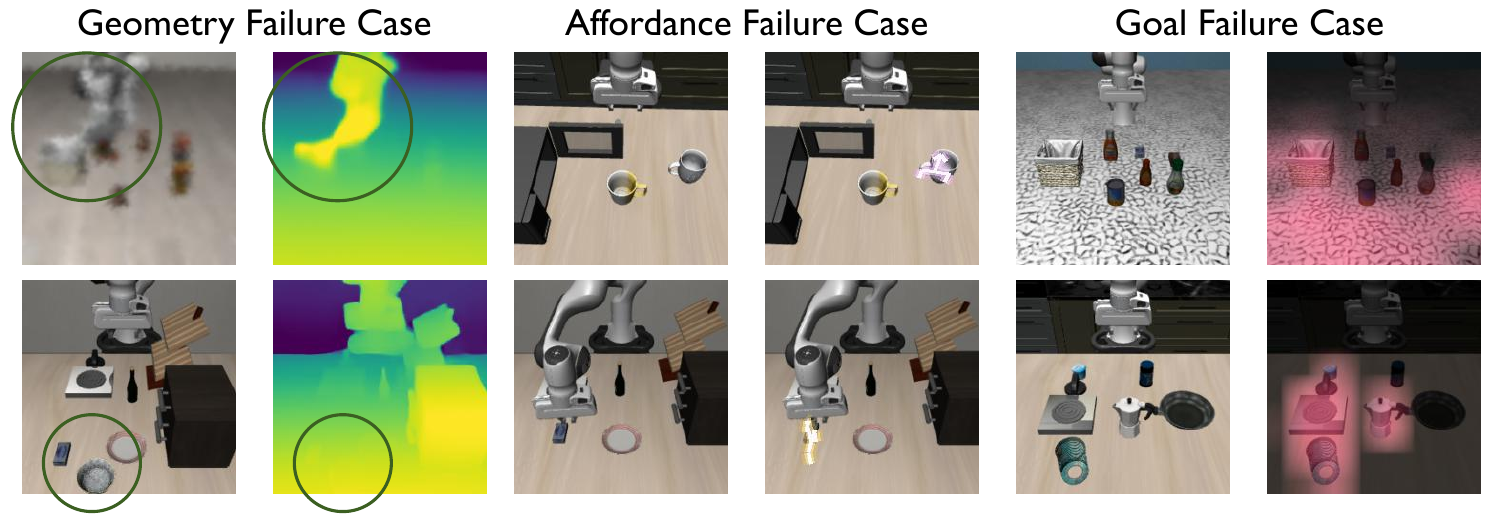}
\caption{\textbf{Failure cases of structured intermediate guidance.}
Geometry can degrade under sensor noise or subtle depth variations;
affordance may confuse task-relevant and distractor objects;
and goal prediction may include task-irrelevant regions under out-of-distribution visual changes.
Colors in the affordance predictions distinguish anchor and other objects.}
\label{fig:failure}
\end{figure}

\textbf{Q2: Must auxiliary predictions be injected into the action head?}
Table~\ref{tab:injection_ablation} compares explicit injection with using
the auxiliary objectives only to supervise intermediate features.
Removing injection improves GIFT-VLA from 76.7\% to 79.6\% on LIBERO-Plus
and from 58.6\% to 61.4\% on RoboCasa. The same trend holds for both WAM variants,
although the differences are smaller: 0.3/0.2 points for GIFT-WAM-Fast and 0.2/0.3
points for GIFT-WAM-IDM on LIBERO-Plus/RoboCasa. These results show that the observed gains
do not require an additional inference pathway that feeds auxiliary predictions
to the action head. Supervising the intermediate tokens is sufficient; the
auxiliary predictions need not be consumed by the model-specific action generators.

\textbf{Q3: What information is encoded by the supervised intermediate tokens?}
Figure~\ref{fig:guidance} provides a representation-level diagnosis
complementary to the quantitative ablation,
examining whether the supervised intermediate features retain
the physical and task structures required for action generation.
For the geometry visualization, we first reconstruct each student feature
$\tilde g_{t,i}$ using Eq.~\eqref{eq:geometry_reconstruction} and substitute it
for the final-layer VGGT feature supplied to the frozen depth head; the earlier
multi-scale features required by that head remain image-derived VGGT features.
This hybrid decoding is used only as a diagnostic of compatibility with the
teacher's geometric representation: depth is neither a training target nor an
input to the action generator.
Across the VLA and WAM examples,
geometry preserves coherent scene layout and object extent despite substantial
camera viewpoint and background change;
affordance predictions remain anchored to the target objects
and recover plausible end-effector configurations; and goal masks
consistently isolate instruction-relevant objects and destination regions.
Rather than merely enriching visual features, the three signals organize
intermediate features around motion
feasibility, structured interactions, and task relevance. Together with the
factor-specific gains in Table~\ref{tab:ablation_full}, these visualizations support
the view that geometry, affordance, and goal form a complementary control-oriented
representation basis shared by semantics-centered and predictive policies.

\textbf{Q4: How does structured intermediate supervision guide the action-generation process?}
Figures~\ref{fig:attention-liberoplus} and~\ref{fig:attention-robocasa} visualize
the attention from action representations to visual features in matched baseline
and GIFT rollouts. On LIBERO-Plus, GIFT maintains stronger and more consistent
emphasis on the manipulated object, target receptacle, and contact-relevant regions
under unseen perturbations as actions unfold, whereas the corresponding baselines attend more
diffusely or shift toward task-irrelevant scene regions. On RoboCasa, the same
pattern extends to a distinct humanoid embodiment and household environment: GIFT
connects action generation to task-relevant regions throughout the rollout. Together with the paired gains in
Tables~\ref{tab:libero_plus} and~\ref{tab:robocasa}, these examples indicate that
structured intermediate supervision helps the action-generation process draw consistently on
manipulation-relevant visual evidence across both environmental perturbations and
changes in embodiment.

\textbf{Q5: When can structured intermediate guidance fail?}
Figure~\ref{fig:failure} shows that guidance becomes unreliable when an intermediate
prediction no longer preserves the corresponding task-relevant structure. Under severe
sensor noise, geometry may fail to recover a plausible scene structure, leading to
incorrect depth judgments during action generation. It may also miss subtle depth
variations on the surfaces of small manipulated objects, weakening the spatial cues
needed for precise grasping. Affordance failures arise when a task-relevant object is
misidentified and the predicted grasping pose is instead associated with a distractor;
this ambiguity becomes more pronounced under large viewpoint changes. In the upper
example, for the instruction ``\textit{Put the yellow and white mug in the microwave and close
it}'', the intermediate features encode the white mug rather than the intended
yellow-and-white mug. In the lower example, the grasping pose of the cream cheese is
recovered, but the object is assigned to the other-object category, as indicated by the
different colors in the figure. Both errors can redirect action generation toward the
wrong object or interaction role. Goal guidance can likewise fail under challenging background shifts
or when a scene contains many unseen objects. In these out-of-distribution cases, the
predicted mask may extend beyond task-relevant regions to cover background areas or
unseen objects, thereby interfering with correct task execution. These cases show that
structured supervision does not eliminate perceptual ambiguity: its effectiveness
depends on whether the intermediate tokens preserve accurate scene structure,
interaction roles, and task relevance under the encountered distribution shift.

\begin{table}[!t]
\centering
\caption{Effect of inference-time denoising steps on GIFT-WAM.}
\small
\begin{tabular}{l cccc}
\toprule
Denoising Steps & 2 & 5 & 10 & 15 \\
\midrule
GIFT-WAM-IDM & 87.8 & 87.6 & 87.7 & 87.6 \\
\midrule
GIFT-WAM-Fast & 72.6 & 72.5 & 72.6 & 72.3 \\
\bottomrule
\end{tabular}%
\label{tab:wam-design}
\vspace{0.3em}

\caption{Action-head design choices for GIFT-VLA on LIBERO-Plus, comparing an MLP with BridgeAttention and evaluating whether the action head uses action-query features alone or jointly attends to visual features.}
\begin{tabular}{ccc c}
\toprule
\multicolumn{2}{c}{Feature} & \multirow{2}{*}{Bridge} & \multirow{2}{*}{SR (\%)} \\
\cmidrule(lr){1-2}ActionQuery & Vision & & \\
\midrule
\checkmark & & MLP & 75.0 \\
\checkmark & & Attention & 75.4 \\
\checkmark & \checkmark & Attention & 75.7 \\
\bottomrule
\end{tabular}%
\label{tab:vla-design}
\end{table}

\subsection{Design Choices}

\textbf{Denoising steps.}
Table~\ref{tab:wam-design} examines the effect of the number of inference-time
denoising steps on both GIFT-WAM-IDM and GIFT-WAM-Fast on LIBERO-Plus. For
GIFT-WAM-Fast, each setting specifies the number of action-denoising steps; for
GIFT-WAM-IDM, it specifies the number of steps used separately for future-video
and action denoising. Across
2, 5, 10, and 15 steps, GIFT-WAM-IDM varies only from 87.6\% to 87.8\%, while
GIFT-WAM-Fast varies from 72.3\% to 72.6\%. Two steps attain the best result for
GIFT-WAM-IDM and tie the best result for GIFT-WAM-Fast, indicating that both
variants can recover effective predictions with a short denoising trajectory.
We therefore use two denoising steps to reduce inference cost without sacrificing
task performance.

\textbf{Action head.}
Table~\ref{tab:vla-design} isolates the effect of the GIFT-VLA action-head design
on LIBERO-Plus. The first row retains the original StarVLA-OFT formulation, which
maps the action-query features to actions through an MLP and achieves 75.0\%.
Replacing the MLP with cross-attention slightly improves the success rate to
75.4\%. Allowing the action latents to additionally attend to visual features,
which corresponds to the no-injection architecture in
Fig.~\ref{fig:inject}(b) without structured intermediate guidance, further raises
the result to 75.7\%. This 0.7-point total gain is modest compared with the
79.6\% achieved by full GIFT-VLA in Table~\ref{tab:ablation_full}: structured
intermediate guidance contributes an additional 3.9 points over the no-guidance
action-head variant. These results indicate that the main performance improvement
comes from structured intermediate supervision rather than the action-head change.
We adopt the cross-attention design solely to provide a controlled interface for
testing optional auxiliary-feature injection; it is not required by GIFT supervision.

\section{Real-World Experiments}

\subsection{Experimental Setup}
\subsubsection{Hardware Setup}
\begin{figure*}[t]
\centering
\includegraphics[width=\textwidth]{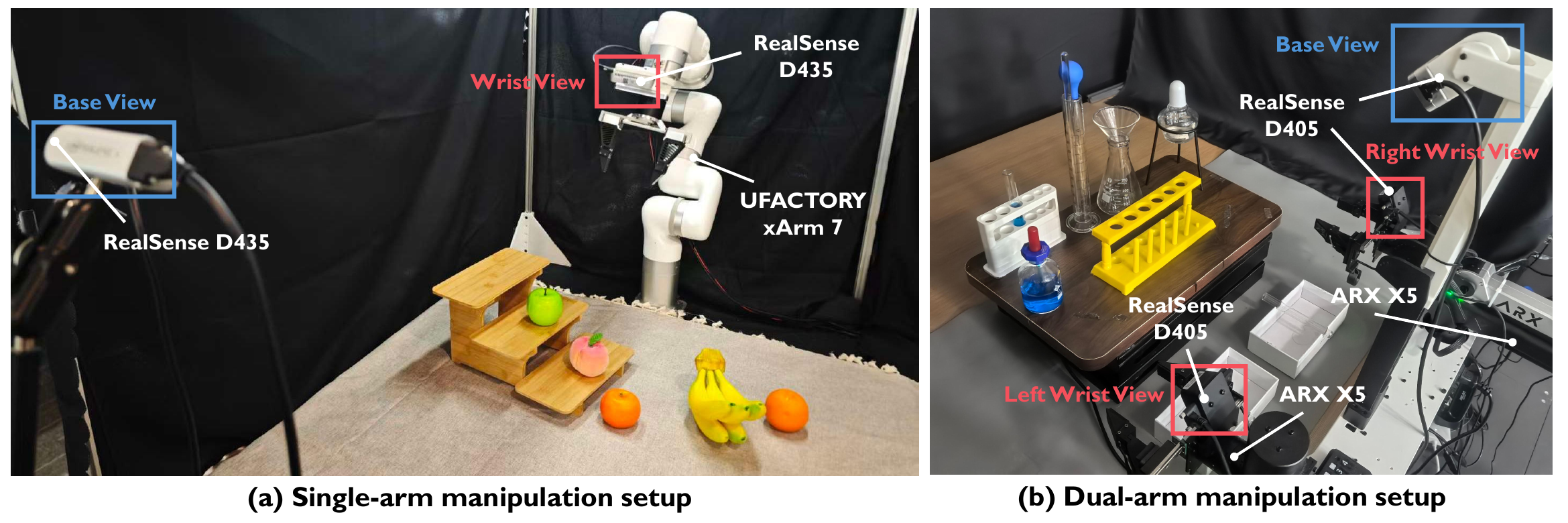}
\caption{\textbf{Real-world robotic platforms.} (a) Single-arm UFACTORY xArm7 setup observed by third-person and wrist-mounted Intel RealSense D435 cameras. (b) Dual-arm ARX X5 setup observed by one head-mounted and two wrist-mounted Intel RealSense D405 cameras. Both systems use parallel grippers.}
\label{fig:real_setup}
\end{figure*}

To evaluate our method in real-world scenarios, we construct the two robotic
manipulation systems shown in Fig.~\ref{fig:real_setup}. Figure~\ref{fig:real_setup}(a)
shows a UFACTORY xArm7 arm observed by two Intel RealSense D435 cameras: one
provides a third-person view and the other is mounted on the wrist, mirroring the
camera configuration used in LIBERO. Figure~\ref{fig:real_setup}(b) shows an ARX
X5 dual-arm platform equipped with three Intel RealSense D405 cameras, including
one head-mounted camera and one wrist-mounted camera for each arm. Both systems
use parallel grippers as their end effectors.

\subsubsection{Data Collection}

As illustrated in Figs.~\ref{fig:singlearm} and~\ref{fig:dualarm}, our
real-world benchmark comprises four tasks spanning the two robotic platforms.
On the xArm7 platform, \textbf{Task 1 (color-conditioned placement)} requires
placing a specified object into the basket of the instructed color, evaluating
semantic understanding, while \textbf{Task 2 (layer-conditioned placement)}
requires placing a specified object on the instructed tier of a display rack,
evaluating spatial reasoning. On the ARX X5 platform, \textbf{Task 3
(articulated-object manipulation)} requires placing a Rubik's cube inside a
cabinet and closing its door, while \textbf{Task 4 (bimanual test-tube
insertion)} requires inserting test tubes into a rack in both upright and
inverted orientations. The latter two tasks evaluate articulated-object interaction and
precise bimanual manipulation, respectively.
We collect 100 human demonstrations per task through
teleoperation, yielding 400 trajectories in total.

During collection, the demonstrator wears green gloves. To reduce the
train--test discrepancy caused by human hands appearing in the recorded
observations, we randomly apply video inpainting to remove the gloves from half
of the collected data and train on a mixture of the inpainted and original
gloved demonstrations.

\subsubsection{Data Annotation} 
We construct real-world object masks and poses through a human-verified annotation
pipeline. Given a text prompt for each task-relevant object, Grounding
DINO~\cite{liu2024groundingdino} first localizes the object, and
SAM2~\cite{ravi2024sam2segmentimage} produces its initial mask. An annotator verifies and, when
necessary, corrects this mask before SAM2 propagates it throughout the complete
episode. We then manually inspect the propagated masks and correct any tracking
errors. Finally, following HumanEgo's use of object-pose
annotations~\cite{wang2026humanego}, we construct a per-frame 6-DoF pose label for
each object. Specifically, we combine each verified mask with the synchronized
RGB-D observation: masked depth determines the object position, while Orient
Anything V2~\cite{wang2025orientanythingv2} estimates its orientation. We annotate
every task-relevant object mentioned in the
language instruction and select the first object interacted with in each
demonstration as the anchor. Together with the poses and closure states of both
end effectors on the ARX platform, these annotations provide temporally consistent
targets for goal and affordance supervision. For bimanual tasks, the instruction
specifies the operation assigned to each arm. Each object is therefore paired
with its designated left or right arm, and its interaction field records the
corresponding end-effector pose in the object frame. We set $K=4$ for the
single-arm xArm7 Tasks~1--2 and $K=5$ for the dual-arm ARX X5 Tasks~3--4.
All allocated slots are supervised without padding or slot masking. In the
bimanual test-tube task, the five slots correspond to the two end effectors and
three instruction-relevant objects.

\begin{table}[t]
\caption{Real-world task performance under the original settings.}
\label{tab:real-world-main}
\centering
\footnotesize
\setlength{\tabcolsep}{3pt}
\begin{tabular}{lccccc}
\toprule
Method & Task 1 & Task 2 & Task 3 & Task 4 & Avg. (\%) \\
\midrule
StarVLA-OFT~\cite{community2026starvla} & 6/10 & 2/10 & 5/10 & 1/10 & 35.0 \\
\textbf{GIFT-VLA} & 7/10 & 5/10 & 7/10 & 4/10 & 57.5 \\
Fast-WAM-IDM~\cite{yuan2026fast} & 7/10 & 4/10 & 8/10 & 2/10 & 52.5 \\
\textbf{GIFT-WAM-IDM} & \textbf{9/10} & \textbf{8/10} & \textbf{10/10} & \textbf{8/10} & \textbf{87.5} \\
\bottomrule
\end{tabular}
\end{table}

\subsubsection{Implementation Details}
We evaluate GIFT-VLA and GIFT-WAM-IDM against their matched baselines,
StarVLA-OFT and Fast-WAM-IDM, respectively. The GIFT model and baseline use
the same training configuration within each policy family. Models for the
single-arm and dual-arm platforms are trained separately. All models are
trained on the collected real-world demonstrations for 50,000 iterations using
64 Zhenwu 810E PPUs, each with 96~GB of device memory. The per-device
batch size is 4 for WAM-based models and 2 for VLA-based models. In all
real-world experiments, actions are represented as relative joint
commands for both training and execution. Both VLA- and WAM-based policies
predict 32-step action chunks and replan every 32 control steps.

\begin{table}[t]
\caption{Real-world robustness under progressively stronger perturbations.
For Task~2, Level~1 changes the illumination and Level~2 additionally replaces
the tabletop background; for Task~4, Level~1 replaces the background objects and
Level~2 additionally rotates the task-relevant objects by approximately
$15^\circ$--$20^\circ$.}
\label{tab:real-world-robustness}
\centering
\footnotesize
\begin{tabular}{lcccc}
\toprule
\multirow{2}{*}{Method} & \multicolumn{2}{c}{Task 2} & \multicolumn{2}{c}{Task 4} \\
\cmidrule(lr){2-3}\cmidrule(lr){4-5}
& Level 1 & Level 2 & Level 1 & Level 2 \\
\midrule
StarVLA-OFT~\cite{community2026starvla} & 1/10 & 1/10 & 0/10 & 0/10 \\
\textbf{GIFT-VLA} & 3/10 & 3/10 & 4/10 & 1/10 \\
Fast-WAM-IDM~\cite{yuan2026fast} & 3/10 & 2/10 & 1/10 & 0/10 \\
\textbf{GIFT-WAM-IDM} & \textbf{7/10} & \textbf{7/10} & \textbf{8/10} & \textbf{5/10} \\
\bottomrule
\end{tabular}
\end{table}

\subsection{Task Setting}
\begin{figure*}[t]
\centering
\includegraphics[width=\textwidth]{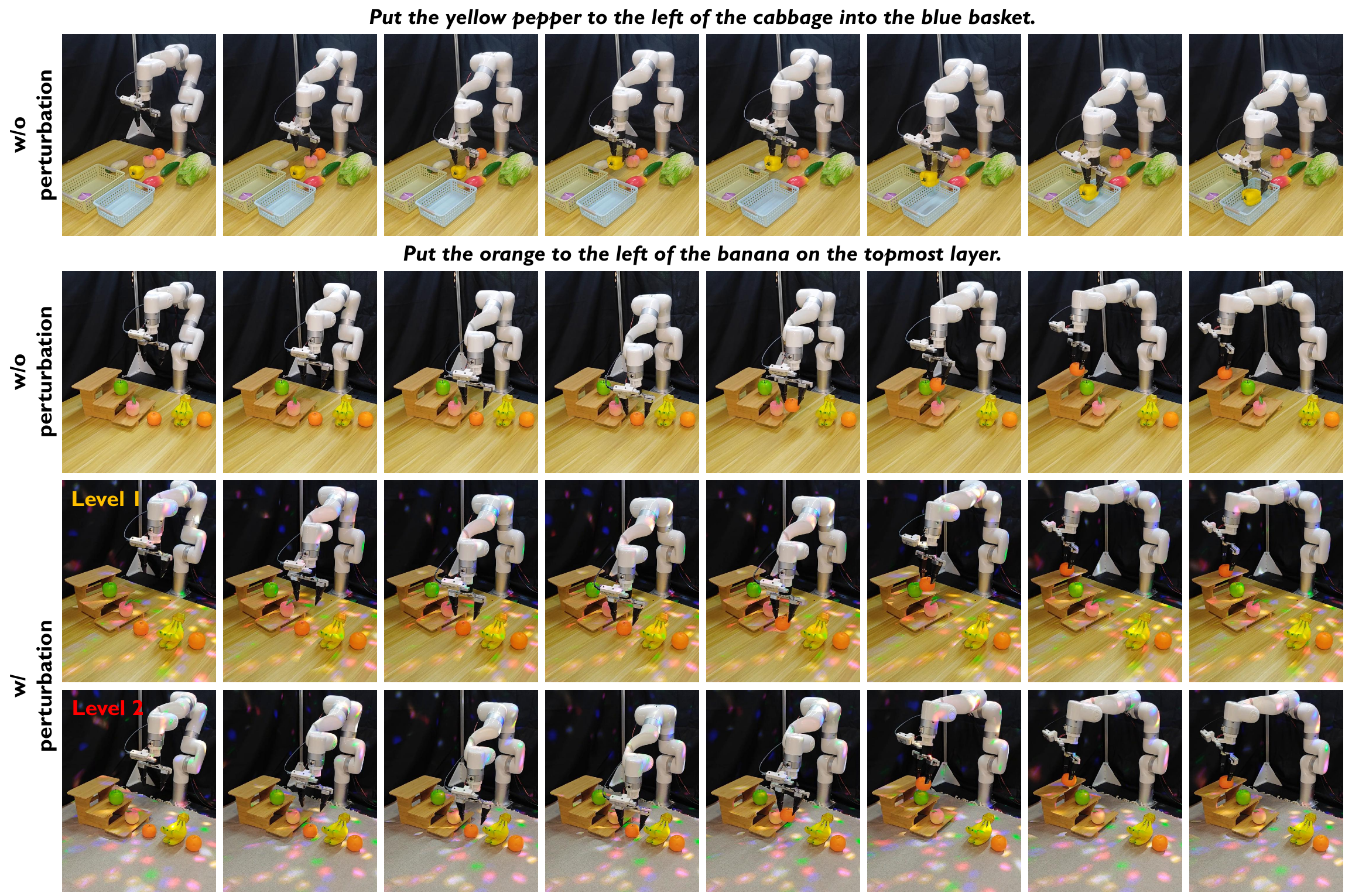}
\caption{\textbf{Visualization of real-world manipulation rollouts on the single-arm platform.} For robustness evaluation, Level 1 introduces a rotating colored lamp, while Level 2 additionally replaces the tabletop background with a tablecloth of a different color.}
\label{fig:singlearm}
\end{figure*}

We evaluate all four tasks under their original settings using 10 trials per
task and method. To further assess robustness to distribution shifts, we perturb
one representative task on each platform: Task~2 (layer-conditioned placement)
on the single-arm xArm7 and Task~4 (bimanual test-tube insertion) on the dual-arm
ARX X5. Following the perturbation categories in LIBERO-Plus~\cite{fei2025libero},
we introduce changes in lighting, background appearance, background objects, and
task-relevant object poses, which also represent common sources of variation in
real-world deployment. Rather than applying synthetic photometric transformations
with preset brightness or contrast ranges, Task~2 Level~1 introduces a rotating
colored lamp positioned above the robot arm so that its illumination covers the
entire tabletop. The lamp produces time-varying colored
illumination, and Level~2 additionally replaces the original tabletop background
with a tablecloth of a different color. For Task~4, Level~1 replaces all
background objects with novel objects, while Level~2 further rotates both the
target test tubes and the test-tube rack by approximately $15^\circ$--$20^\circ$
relative to their training configurations. None of these lighting, background-object,
tablecloth, or task-relevant pose conditions appears in the training data. Each
perturbation level is evaluated over 10 trials per method. As visualized in
Figs.~\ref{fig:singlearm} and~\ref{fig:dualarm}, these evaluations test whether
the structured intermediate features learned by GIFT remain reliable
under substantial visual and spatial variations.

\subsection{Results}

\textbf{Quantitative results.}
Table~\ref{tab:real-world-main} compares the four methods under the original
task settings. GIFT improves both policy families consistently across all four
tasks: GIFT-VLA raises the overall success rate from 35.0\% for StarVLA-OFT to
57.5\%, while GIFT-WAM-IDM improves Fast-WAM-IDM from 52.5\% to 87.5\%.
The gains span color-conditioned placement, layer-conditioned placement,
articulated-object manipulation, and bimanual test-tube insertion. In particular,
GIFT-WAM-IDM achieves 8/10 on Task~4 compared with 2/10 for Fast-WAM-IDM,
consistent with the benefit of explicitly supervising interaction roles
and object-centric end-effector configurations for precise
bimanual interaction.

\begin{figure*}[!t]
\centering
\includegraphics[width=\textwidth]{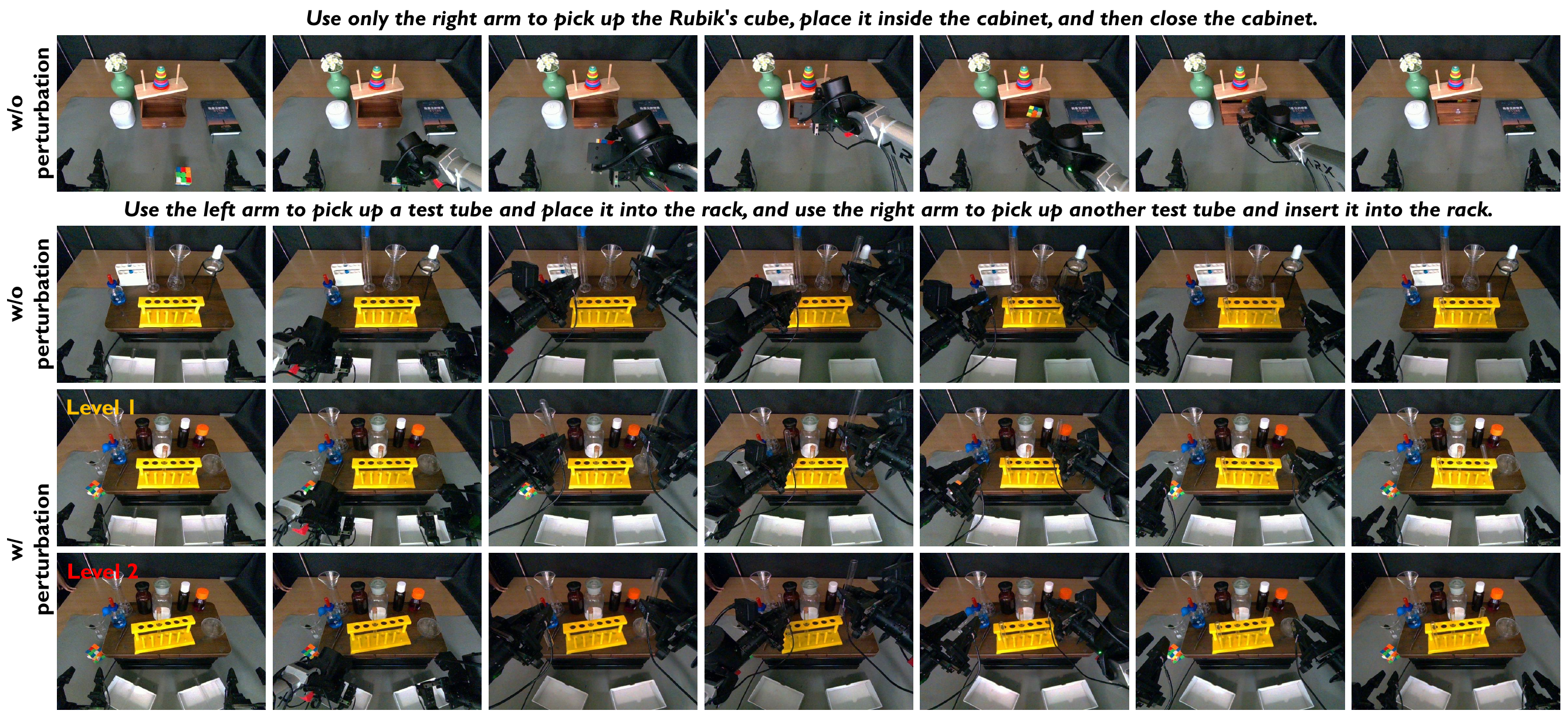}
\caption{\textbf{Visualization of real-world manipulation rollouts on the dual-arm platform.} Level 1 replaces all background objects with novel objects, while Level 2 additionally rotates the target test tubes and rack by approximately $15^\circ$--$20^\circ$ relative to their training configurations.}
\label{fig:dualarm}
\end{figure*}

Table~\ref{tab:real-world-robustness} further evaluates robustness under unseen
appearance and spatial perturbations. Aggregated over the four perturbation
conditions, GIFT-VLA and GIFT-WAM-IDM achieve success rates of 27.5\% and
67.5\%, substantially exceeding StarVLA-OFT at 5.0\% and Fast-WAM-IDM at
15.0\%, respectively. On Task~2, GIFT-WAM-IDM maintains 7/10 success under both
perturbation levels. On the more challenging Task~4, it achieves 8/10 under
Level~1 and 5/10 under Level~2. These results show that joint supervision of
geometry, affordance, and goals reduces reliance on background and layout
shortcuts, yielding
control-oriented intermediate features across both VLA and WAM
policies without requiring auxiliary predictions as action inputs.
The larger margins over the matched baselines in the real-world experiments than
in simulation are concentrated on Tasks~2 and~4, both of which impose tight
control tolerances. Task~2 requires accurately localizing the instructed rack tier
and placing the object within a narrow valid region, whereas Task~4 demands
high-precision bimanual insertion in both upright and inverted orientations.
These requirements become more stringent under perturbations, where success
depends on robust estimates of spatial relations and interaction poses. Explicit
geometry and affordance supervision therefore provide a larger advantage in
these precision-sensitive settings, helping explain the enlarged real-world
margins. Overall, these results highlight the particular value of structured
guidance for precision-sensitive manipulation under substantial perturbations.

\textbf{Qualitative results.}
Under unperturbed conditions, GIFT reliably executes the tasks on both real-world
platforms, as shown in Figs.~\ref{fig:singlearm} and~\ref{fig:dualarm}. On the
single-arm xArm7, it completes \textbf{Task 1} by placing the pepper into the
blue basket and \textbf{Task 2} by placing the orange on the topmost tier. On
the dual-arm ARX X5, it completes \textbf{Task 3} by placing the Rubik's cube
inside the cabinet and closing its door. In \textbf{Task 4}, it performs
fine-grained bimanual coordination to grasp and accurately insert separate test
tubes into the rack in upright and inverted orientations. More importantly,
GIFT retains higher robustness than its matched baselines as previously unseen
perturbations become progressively stronger, with the advantage being most pronounced
for GIFT-WAM-IDM. For the layer-placement task, we first change
the lighting conditions and then compound this appearance shift by also changing the
background. For the test-tube
insertion task, we first rearrange background objects and then
further change the poses of task-relevant objects. The successful GIFT rollouts
shown here preserve the required target selection, spatial relations, and precise
control under these compound appearance and spatial shifts. Together with the
quantitative results in Table~\ref{tab:real-world-robustness}, these examples show
that GIFT, particularly GIFT-WAM-IDM, generalizes more reliably than its matched
baselines beyond the unperturbed collection conditions.

\section{Conclusion}
We presented GIFT, an architecture-flexible framework that turns complementary
physical and task structures into explicit supervision for the intermediate
features of robot policies. The GIFT objectives do not require modifying the
model-specific action-generation interface or supplying auxiliary predictions as
action inputs; instead, geometry, affordance, and goal objectives teach
features what information to preserve: scene structure for feasible
motion, object-centric interaction configurations for manipulation, and
instruction-relevant regions for task completion. We instantiated the same
principle in a semantics-centered VLA, a fast-mode WAM, and an inverse-dynamics
WAM, covering distinct backbones and action formulations. On LIBERO and
RoboCasa, GIFT maintained competitive in-distribution control, with especially
large gains on articulated-object interaction. Under zero-shot transfer to
LIBERO-Plus, it consistently improved robustness across distribution shifts,
including changes in robot initialization. Guidance
ablations confirmed that all three signals are complementary, while injection
ablations showed that explicitly feeding auxiliary predictions to the action head
is unnecessary: shaping the shared intermediate features alone is sufficient and can
perform better. Real-robot experiments on single- and dual-arm platforms further
demonstrated robustness to lighting, background, layout, and object-pose changes.
These findings position structured intermediate supervision as a reusable
interface between general-purpose visual knowledge and robot control. Future work
will explore additional forms of guidance for intermediate features, scale up
training, and further improve the reliability of guidance signals under severe
perceptual distribution shifts.

\bibliographystyle{IEEEtran}
\bibliography{reference}

\end{document}